\pdfoutput=1

\documentclass[11pt]{article}

\usepackage{acl}
\usepackage{times}
\usepackage{mathtools}
\usepackage{tikz}
\usepackage{setspace}
\usepackage{latexsym}
\usepackage[T1]{fontenc}
\usepackage[utf8]{inputenc}
\usepackage{array}
\usepackage{hyperref}       
\usepackage{url}            
\usepackage{booktabs}       
\usepackage{amsfonts}       
\usepackage{nicefrac}       
\usepackage{xcolor}         
\usepackage{booktabs} 
\usepackage{fontawesome5}
\usepackage{kotex}
\usepackage{natbib}
\usepackage{arydshln}
\usepackage{adjustbox}
\usepackage{amsmath}
\usepackage{lscape}
\usepackage{multirow}
\usepackage{microtype}
\usepackage{inconsolata}
\usepackage{graphicx}
\usepackage{enumitem}
\usepackage{makecell}
\usepackage{tabularx}
\usepackage{pgfplots}
\usepackage{pgfplotstable}
\usepackage{subcaption}
\usepackage{xcolor}
\usepackage{tikz}
\usepackage{pgfplots}
\usepackage{booktabs}
\usepackage{tcolorbox}
\usepackage{xcolor}
\usepackage{graphicx}
\usepackage{algorithm}
\usepackage{tcolorbox}
\usepackage{tabularx}
\usepackage{colortbl}
\usepackage{xcolor}
\usepackage{algpseudocode}
\usepackage{pgfplots}
\usepackage{pgfplotstable}
\usepgfplotslibrary{groupplots}
\pgfplotsset{compat=1.17}
\usepgfplotslibrary{polar}
\usepackage{tikz}
\usepackage{longtable}
\usetikzlibrary{shapes.geometric}
\usetikzlibrary{shapes.misc} 
\usepackage{amssymb}        
\usetikzlibrary{positioning, arrows.meta, shapes.multipart, shadows, calc}

\pgfplotsset{compat=1.17}

\usepgfplotslibrary{groupplots}
\usetikzlibrary{patterns}
\pgfplotsset{compat=1.17}

\title{RCScore: Quantifying Response Consistency in Large Language Models}
\author{
  Dongjun Jang \and Youngchae Ahn \and Hyopil Shin \\
  Department of Linguistics \\
  Seoul National University \\
  \texttt{\{qwer4107, estelle1026, hpshin\}@snu.ac.kr}
}

\usepackage[T1]{fontenc}  
\usepackage[utf8]{inputenc}  
\usepackage{tipa}  

\begin{document}
\maketitle

\begin{abstract}
Current LLM evaluations often rely on a single instruction template, overlooking models' sensitivity to instruction style—a critical aspect for real-world deployments. We present RCScore, a multi-dimensional framework quantifying how instruction formulation affects model responses. By systematically transforming benchmark problems into multiple instruction styles, RCScore reveals performance variations undetected by conventional metrics. Our experiments across ten LLMs on four reasoning benchmarks demonstrate that instruction style can shift accuracy by up to 16.7\% points. We introduce Cross-Response Similarity (CRS), a method applying RCScore metrics to measure stylistic self-consistency, and establish its strong correlation with task accuracy, suggesting consistency as a valuable proxy for model reliability. Additional findings show that deterministic decoding produces more stylistically stable outputs, and model scale correlates positively with cross-style consistency. RCScore offers a principled approach to assess instruction robustness.
\end{abstract}

\section{Introduction}

Large language models (LLMs) exhibit remarkable proficiency in complex reasoning, multi-step problem-solving, and creative generation, significantly extending their utility beyond fundamental text generation \cite{brown2020language, touvron2023llama}. While these models demonstrate advanced cognitive capabilities in areas such as mathematical reasoning, logical deduction, and code generation, current evaluation benchmarks inadequately capture the subtleties of these sophisticated tasks.

Current evaluation practices for LLMs predominantly rely on traditional metrics like accuracy and F1 scores. Although mathematically validated, these metrics fall short in assessing the quality, consistency, and multifaceted aspects of open-ended LLM responses, where diverse valid outputs are common. Despite impressive scores on reasoning-intensive benchmarks like MMLU \cite{hendrycks2020measuring}, GPQA \cite{rein2024gpqa}, and GSM8K \citep{cobbe2021training}, this reliance creates a fundamental disconnect between the sophisticated outputs of LLMs and the simplistic nature of their evaluation.

To address these limitations, one strategy that has been explored is the use of LLMs as evaluators \citep{zheng2023judging, xia2025evaluating}. While offering flexibility and scalability, this approach faces challenges including domain knowledge constraints requiring costly fine-tuning and inherent model biases that potentially undermine evaluation objectivity. Critically, it introduces a circular evaluation paradigm, where LLMs assess other LLMs, compromising objectivity and reliability as identified by \citet{wang2023large}.

A particularly significant yet underexplored evaluation gap concerns the impact of instruction formulation on model performance. Research has demonstrated that LLM performance varies considerably depending on prompt design and instruction style \citep{liu2023pre}. Beyond methods such as Chain-of-Thought \citep{wei2022chain} and Tree-of-Thoughts \citep{yao2023tree}, recent studies have highlighted the sensitivity of language models to instruction phrasing. \citet{ajith2023instructeval} and \citet{cao2024worst} analyze performance variation under prompt perturbations, while \citet{mizrahi2024state} show that evaluations based on a single template can yield unreliable metrics across paraphrased instructions. This sensitivity extends to subtle linguistic variations as shown by \citet{wahle2024paraphrase} and \citet{leidinger2023language}. Following this, recent studies propose metrics for prompt sensitivity, including rephrasing-based evaluation \citep{lu2024prompts, errica2024did} and embedding-based coherence scoring \citep{lauriola2025analyzing}. However, these approaches rely on ad-hoc rewriting, incur high computational costs, and offer limited insight into model behavior. 

To address this evaluation gap, we introduce RCScore\footnote{RCScore github link \url{https://github.com/Junmaij/RCScore}}, a comprehensive metric that methodically assesses how LLMs respond to different instruction styles. Through structured variation of instructional cues across clause types, RCScore evaluates responses across three key dimensions—Structurality, Lexicality, and Coherence—providing deeper insights into model capabilities. Our metric delivers three important contributions: (1) it reveals performance variations across instruction styles that traditional accuracy-focused benchmarks often miss; (2) it introduces Cross-Response Similarity (CRS) to measure stylistic self-consistency, showing that higher consistency correlates with better task accuracy, suggesting consistency as a reliability indicator; and (3) it provides an easy-to-reproduce, efficient protocol for assessing model robustness.

\section{Related Work}
\subsection{Traditional LLM Evaluation Methods}
As LLMs have advanced rapidly, a variety of evaluation methodologies have emerged to assess their capabilities. However, most mainstream benchmarks—such as MMLU \citep{hendrycks2020measuring}, BIG-Bench \citep{srivastava2022beyond}, GSM8K \citep{cobbe2021training}, GPQA \citep{rein2024gpqa}, HellaSwag \citep{zellers2019hellaswag}, MATH \citep{hendrycks2021measuring}, and DROP \citep{dua2019drop}—continue to rely on traditional metrics such as exact match accuracy and F1 score. This simplification is further reflected in the technical reports of many recent LLMs, which predominantly emphasize accuracy-based performance results \citep{achiam2023gpt, grattafiori2024llama, team2023gemini, yang2024qwen2, team2025gemma, claude3, guo2025deepseek}. \citet{mondorf2024beyond} argue that LLMs’ reasoning abilities remain unclear due to an overreliance on shallow accuracy-based metrics. Other studies similarly critique evaluations based solely on final answers, and propose alternative methods  \citep{mahdavi2025brains, nguyen2024direct, golovneva2023roscoe}.

\subsection{Beyond Accuracy-Centric Evaluation}
To address the limitations of accuracy-centric evaluation, recent studies have explored alternative strategies that go beyond exact match or lexical overlap. One prominent direction is the \textit{LLM-as-a-judge} paradigm, where strong models are used to assess the quality of generated responses directly \citep{zheng2023judging, xia2025evaluating}. Yet, studies show that such methods can diverge from human judgments and suffer from structural flaws \citep{wang2023large, wang2025can}, casting doubt on their reliability. In parallel, other studies assess different facets of LLM output quality. FactScore \citep{min2023factscore} verifies generations by checking atomic factual units against external sources. SelfCheckGPT \citep{manakul2023selfcheckgpt} evaluates factuality through consistency across sampled outputs, and ARES \citep{saad2023ares} scores the quality of individual components in retrieval-augmented generation pipelines.

\subsection{Stylistic Variation in Prompting: Effects and Evaluation}
Prompt formulation plays a critical role in shaping LLM behavior and performance, often enabling substantial gains without parameter updates \citep{liu2023pre}. The field evolved from zero-shot \citep{radford2019language} and few-shot approaches \citep{brown2020language} to more sophisticated techniques like Chain-of-Thought \citep{wei2022chain}, Tree of Thoughts \citep{yao2023tree}, self-refine prompting \citep{madaan2023self}, and Automatic Prompt Engineer \citep{zhou2023large}, demonstrating prompt design's significant impact on LLM capabilities. 

As prompting techniques developed, research shifted toward understanding how stylistic variations across prompts affect model performance. \citet{he2024does} demonstrated that surface-level formatting changes (e.g., JSON, YAML) affect model outputs. On the linguistic side, \citet{wahle2024paraphrase} examined the impact of morphological, syntactic, and lexical paraphrases, while \citet{leidinger2023language} highlighted LLMs' sensitivity to subtle linguistic variations. Using prompt variants that are semantically equivalent, these studies demonstrated that models are highly sensitive to even subtle stylistic cues. 

Motivated by these findings, several studies have sought to formalize the effect of prompt variability by designing dedicated evaluation metrics. \citet{lu2024prompts} proposed sensitivity-based metrics to assess how prompt formulation influences model performance. \citet{errica2024did} introduced sensitivity and consistency metrics, using paraphrased benchmark questions generated via LLaMA3. \citet{lauriola2025analyzing} focused on coherence, measuring the average cosine similarity between answers to equivalent questions.

\section{Experiment on Instruction Style Sensitivity}
Large language models often exhibit varying performance depending on how instructions are phrased, yet traditional benchmarks rarely account for this variability. Prior work shows that subtle prompt variations affect how models respond \citep{he2024does, wahle2024paraphrase, leidinger2023language}, and \citet{mizrahi2024state} further demonstrate that evaluations relying on a single instruction template are inadequate. Building on these findings, we conducted comprehensive evaluations across diverse reasoning tasks and multiple model families to quantify how different instruction formulations influence model accuracy and response characteristics.

\subsection{Benchmark}
We selected four benchmarks covering diverse reasoning tasks in mathematical and scientific domains. The American Invitational Mathematics Examination (AIME 2024)\footnote{\url{https://huggingface.co/datasets/HuggingFaceH4/aime_2024}} contains competitive mathematics problems that require precise numerical reasoning. GSM8K \citep{cobbe2021training} includes diverse grade school math word problems designed to test multi-step arithmetic reasoning. MATH-500 is a curated subset of 500 problems from the MATH benchmark \citep{hendrycks2021measuring}, selected by OpenAI as part of the \textit{Let's Verify Step by Step} study \citep{lightman2023let}. For advanced scientific reasoning, we use the GPQA-Diamond subset of the GPQA \citep{rein2024gpqa}, which is originally a multiple-choice benchmark but is evaluated in our setting without answer options to better assess pure reasoning ability. These benchmarks were chosen due to their prominent role in recent LLM technical reports \citep{achiam2023gpt, grattafiori2024llama, team2023gemini, yang2024qwen2, team2025gemma, claude3, guo2025deepseek}.

\subsection{Models}
Our evaluation covers ten open-source instruction-tuned LLMs spanning three major model families. This includes the Gemma 3 series from Google—Gemma 3 4B, 12B, and 27B Instruct variants \citep{team2025gemma}; the Qwen 2.5 Instruct models from Alibaba at 3B, 7B, 32B, and 72B scales \citep{yang2024qwen2}; and Meta's Llama 3 family—Llama 3.2 3B, Llama 3.1 8B, and Llama 3.3 70B Instruct models \citep{grattafiori2024llama}. 

\subsection{Instruction Style Variations}
\begin{figure}[!h]
\centering
\begin{tcolorbox}[
    width=0.49\textwidth,  
    colback=blue!5,
    colframe=black!40,
    boxrule=0.5pt,
    arc=1mm,
    fontupper=\footnotesize,  
    left=4pt,  
    right=4pt,
    top=5pt,
    bottom=5pt,
    boxsep=3pt,
]
\begin{tabularx}{\textwidth}{>{\raggedright\arraybackslash\hsize=0.20\hsize}X>{\raggedright\arraybackslash\hsize=0.80\hsize}X}
\multicolumn{2}{p{\linewidth}}{\raggedright\textbf{\textcolor{black!80}{Instruction style variations for Style Sensitivity Experiment}}} \\
\midrule
\rowcolor{blue!10} \textbf{Style} & \textbf{Characteristics and Example} \\
\midrule
\textbf{\textcolor{blue!70!black}{Declarative}} & \textit{Statements presenting facts or information} \\
& "The problem should be solved step by step. The answer is to be suggested in the following format." \\
\midrule
\textbf{\textcolor{violet!80!black}{Interrogative}} & \textit{Direct questions challenging reasoning} \\
& "Could you solve the problem step by step? Would you suggest the answer in the following format?" \\
\midrule
\textbf{\textcolor{green!60!black}{Exclamative}} & \textit{Expressions emphasizing importance or difficulty} \\
& "How important it is to solve the problem step by step! What a necessity it is to suggest the answer in the following format!" \\
\midrule
\textbf{\textcolor{orange!80!black}{Imperative}} & \textit{Command-based instructions directing processes} \\
& "Solve the problem step by step. Suggest the answer in the following format." \\
\end{tabularx}
\end{tcolorbox}
\caption{Instruction Styles Employed in Experiments. Detailed prompt templates are provided in Figure \ref{fig_prompt_template}.}
\label{fig:instruction_styles}
\end{figure}

To probe stylistic sensitivity without introducing semantic drift, we adopted a minimal intervention strategy. Rather than altering the benchmark question itself, we applied controlled stylistic variations solely to the instruction prefix. This design enables targeted analysis of how stylistic differences in prompts influence model responses, while ensuring that the core task input remains unchanged. Our typology follows the syntactic clause classification of \citet{huddleston2002cambridge}, comprising Declarative, Interrogative, Exclamative, and Imperative forms. We ensured semantic consistency by fixing main verbs (e.g., solve, suggest) and maintaining comparable lexical complexity across styles (Type Token Ratio: 0.75–0.78). The prompt structure used in our experiments is illustrated in Figure~\ref{fig_prompt_template}.

\begin{figure*}[t]
\centering 

\begin{subfigure}{\textwidth}
    \centering
    \setlength{\tabcolsep}{1pt}
    \begin{tabular}{cccc}
    \begin{tikzpicture}[scale=1.0]
    \begin{axis}[
        title={\small\textbf{AIME}},
        xlabel={\small Style},
        ylabel={\small Accuracy (\%)},
        xmin=0.5, xmax=4.5,
        ymin=0, ymax=35,
        xtick={1,2,3,4},
        xticklabels={\small S1, \small S2, \small S3, \small S4},
        ytick={0,10,20,30},
        yticklabels={\small 0, \small 10, \small 20, \small 30},
        width=4cm,
        height=4.5cm,
        axis line style={draw=black!50},
        tick style={draw=black!50},
        grid=none,
    ]
    \draw[thick, fill=blue!10, draw=black!50] (axis cs:0.75, 3.3) rectangle (axis cs:1.25, 30.0);
    \draw[thick, draw=black!50] (axis cs:0.75, 13.3) -- (axis cs:1.25, 13.3);

    \draw[thick, fill=blue!10, draw=black!50] (axis cs:1.75, 0.0) rectangle (axis cs:2.25, 26.7);
    \draw[thick, draw=black!50] (axis cs:1.75, 10.0) -- (axis cs:2.25, 10.0);

    \draw[thick, fill=blue!10, draw=black!50] (axis cs:2.75, 3.3) rectangle (axis cs:3.25, 33.3);
    \draw[thick, draw=black!50] (axis cs:2.75, 13.3) -- (axis cs:3.25, 13.3);

    \draw[thick, fill=blue!10, draw=black!50] (axis cs:3.75, 3.3) rectangle (axis cs:4.25, 23.3);
    \draw[thick, draw=black!50] (axis cs:3.75, 13.3) -- (axis cs:4.25, 13.3);

    \addplot[only marks, mark=*, mark size=1.5pt, color=green!70!black] coordinates {(1,30.0) (2,26.7) (3,33.3) (4,20.0)}; 
    \addplot[only marks, mark=diamond*, mark size=2pt, color=green!70!black] coordinates {(1,23.3) (2,23.3) (3,23.3) (4,20.0)}; 
    \addplot[only marks, mark=triangle*, mark size=2pt, color=green!70!black] coordinates {(1,6.7) (2,10.0) (3,6.7) (4,3.3)}; 

    \addplot[only marks, mark=square*, mark size=1.5pt, color=blue!70!black] coordinates {(1,20.0) (2,23.3) (3,26.7) (4,23.3)}; 
    \addplot[only marks, mark=otimes, mark size=2pt, color=blue!70!black] coordinates {(1,3.3) (2,10.0) (3,6.7) (4,6.7)}; 
    \addplot[only marks, mark=square, mark size=2pt, color=blue!70!black] coordinates {(1,3.3) (2,0.0) (3,6.7) (4,10.0)}; 

    \addplot[only marks, mark=star, mark size=2pt, color=red!70!black] coordinates {(1,30.0) (2,23.3) (3,13.3) (4,23.3)}; 
    \addplot[only marks, mark=+, mark size=2.5pt, color=red!70!black] coordinates {(1,13.3) (2,13.3) (3,16.7) (4,16.7)}; 
    \addplot[only marks, mark=x, mark size=2.5pt, color=red!70!black] coordinates {(1,13.3) (2,6.7) (3,13.3) (4,13.3)}; 
    \addplot[only marks, mark=|, mark size=2.5pt, color=red!70!black] coordinates {(1,3.3) (2,3.3) (3,3.3) (4,6.7)}; 
    \end{axis}
    \end{tikzpicture}
    &
    \begin{tikzpicture}[scale=1.0]
    \begin{axis}[
        title={\small\textbf{MATH-500}},
        xlabel={\small Style},
        ylabel={},
        xmin=0.5, xmax=4.5,
        ymin=38, ymax=80,
        xtick={1,2,3,4},
        xticklabels={\small S1, \small S2, \small S3, \small S4},
        ytick={40,50,60,70,80},
        yticklabels={\small 40, \small 50, \small 60, \small 70, \small 80},
        width=4cm,
        height=4.5cm,
        axis line style={draw=black!50},
        tick style={draw=black!50},
        grid=none,
    ]
    \draw[thick, fill=blue!10, draw=black!50] (axis cs:0.75, 41.8) rectangle (axis cs:1.25, 75.8);
    \draw[thick, draw=black!50] (axis cs:0.75, 65.0) -- (axis cs:1.25, 65.0);

    \draw[thick, fill=blue!10, draw=black!50] (axis cs:1.75, 41.6) rectangle (axis cs:2.25, 77.6);
    \draw[thick, draw=black!50] (axis cs:1.75, 64.0) -- (axis cs:2.25, 64.0);

    \draw[thick, fill=blue!10, draw=black!50] (axis cs:2.75, 41.0) rectangle (axis cs:3.25, 76.8);
    \draw[thick, draw=black!50] (axis cs:2.75, 64.0) -- (axis cs:3.25, 64.0);

    \draw[thick, fill=blue!10, draw=black!50] (axis cs:3.75, 39.8) rectangle (axis cs:4.25, 77.4);
    \draw[thick, draw=black!50] (axis cs:3.75, 67.0) -- (axis cs:4.25, 67.0);

    \addplot[only marks, mark=*, mark size=1.5pt, color=green!70!black] coordinates {(1,75.8) (2,77.6) (3,76.8) (4,77.4)}; 
    \addplot[only marks, mark=diamond*, mark size=2pt, color=green!70!black] coordinates {(1,71.4) (2,71.6) (3,73.0) (4,72.4)}; 
    \addplot[only marks, mark=triangle*, mark size=2pt, color=green!70!black] coordinates {(1,65.8) (2,64.0) (3,66.0) (4,67.0)}; 

    \addplot[only marks, mark=square*, mark size=1.5pt, color=blue!70!black] coordinates {(1,64.4) (2,64.4) (3,61.6) (4,63.6)}; 
    \addplot[only marks, mark=otimes, mark size=2pt, color=blue!70!black] coordinates {(1,42.8) (2,43.6) (3,42.4) (4,46.4)}; 
    \addplot[only marks, mark=square, mark size=2pt, color=blue!70!black]  coordinates {(1,41.8) (2,41.6) (3,41.0) (4,39.8)}; 

    \addplot[only marks, mark=star, mark size=2pt, color=red!70!black] coordinates {(1,70.8) (2,69.6) (3,71.2) (4,71.2)}; 
    \addplot[only marks, mark=+, mark size=2.5pt, color=red!70!black] coordinates {(1,69.8) (2,68.8) (3,67.8) (4,69.8)}; 
    \addplot[only marks, mark=x, mark size=2.5pt, color=red!70!black] coordinates {(1,64.0) (2,63.0) (3,58.2) (4,65.0)}; 
    \addplot[only marks, mark=|, mark size=2.5pt, color=red!70!black] coordinates {(1,55.0) (2,54.2) (3,55.2) (4,55.4)}; 
    \end{axis}
    \end{tikzpicture}
    &
    \begin{tikzpicture}[scale=1.0]
    \begin{axis}[
        title={\small\textbf{GPQA-Diamond}},
        xlabel={\small Style},
        ylabel={},
        xmin=0.5, xmax=4.5,
        ymin=1.5, ymax=9,
        xtick={1,2,3,4},
        xticklabels={\small S1, \small S2, \small S3, \small S4},
        ytick={2,4,6,8},
        yticklabels={\small 2, \small 4, \small 6, \small 8},
        width=4cm,
        height=4.5cm,
        axis line style={draw=black!50},
        tick style={draw=black!50},
        grid=none,
    ]
    \draw[thick, fill=blue!10, draw=black!50] (axis cs:0.75, 2.0) rectangle (axis cs:1.25, 6.1);
    \draw[thick, draw=black!50] (axis cs:0.75, 4.5) -- (axis cs:1.25, 4.5);

    \draw[thick, fill=blue!10, draw=black!50] (axis cs:1.75, 2.0) rectangle (axis cs:2.25, 8.6);
    \draw[thick, draw=black!50] (axis cs:1.75, 3.5) -- (axis cs:2.25, 3.5);

    \draw[thick, fill=blue!10, draw=black!50] (axis cs:2.75, 3.0) rectangle (axis cs:3.25, 6.6);
    \draw[thick, draw=black!50] (axis cs:2.75, 4.5) -- (axis cs:3.25, 4.5);

    \draw[thick, fill=blue!10, draw=black!50] (axis cs:3.75, 2.0) rectangle (axis cs:4.25, 6.6);
    \draw[thick, draw=black!50] (axis cs:3.75, 4.0) -- (axis cs:4.25, 4.0);

    \addplot[only marks, mark=*, mark size=1.5pt, color=green!70!black] coordinates {(1,6.1) (2,7.1) (3,6.1) (4,6.6)}; 
    \addplot[only marks, mark=diamond*, mark size=2pt, color=green!70!black] coordinates {(1,4.5) (2,4.0) (3,6.6) (4,6.6)}; 
    \addplot[only marks, mark=triangle*, mark size=2pt, color=green!70!black] coordinates {(1,4.5) (2,3.0) (3,4.0) (4,2.5)}; 

    \addplot[only marks, mark=square*, mark size=1.5pt, color=blue!70!black] coordinates {(1,6.1) (2,8.6) (3,5.6) (4,5.6)}; 
    \addplot[only marks, mark=otimes, mark size=2pt, color=blue!70!black] coordinates {(1,2.0) (2,3.5) (3,4.5) (4,2.5)}; 
    \addplot[only marks, mark=square, mark size=2pt, color=blue!70!black] coordinates {(1,5.1) (2,2.0) (3,4.0) (4,5.1)}; 

    \addplot[only marks, mark=star, mark size=2pt, color=red!70!black] coordinates {(1,4.0) (2,6.6) (3,5.6) (4,5.1)}; 
    \addplot[only marks, mark=+, mark size=2.5pt, color=red!70!black] coordinates {(1,3.5) (2,2.0) (3,6.1) (4,2.0)}; 
    \addplot[only marks, mark=x, mark size=2.5pt, color=red!70!black] coordinates {(1,5.6) (2,3.5) (3,4.5) (4,4.0)}; 
    \addplot[only marks, mark=|, mark size=2.5pt, color=red!70!black] coordinates {(1,2.5) (2,2.5) (3,3.0) (4,3.5)}; 
    \end{axis}
    \end{tikzpicture}
    &
    \begin{tikzpicture}[scale=1.0]
    \begin{axis}[
        title={\small\textbf{GSM8K}},
        xlabel={\small Style},
        ylabel={},
        xmin=0.5, xmax=4.5,
        ymin=70, ymax=96,
        xtick={1,2,3,4},
        xticklabels={\small S1, \small S2, \small S3, \small S4},
        ytick={70,80,90},
        yticklabels={\small 70, \small 80, \small 90},
        width=4cm,
        height=4.5cm,
        axis line style={draw=black!50},
        tick style={draw=black!50},
        grid=none,
    ]
    \draw[thick, fill=blue!10, draw=black!50] (axis cs:0.75, 78.3) rectangle (axis cs:1.25, 94.8);
    \draw[thick, draw=black!50] (axis cs:0.75, 89.0) -- (axis cs:1.25, 89.0);

    \draw[thick, fill=blue!10, draw=black!50] (axis cs:1.75, 76.4) rectangle (axis cs:2.25, 93.9);
    \draw[thick, draw=black!50] (axis cs:1.75, 89.0) -- (axis cs:2.25, 89.0);

    \draw[thick, fill=blue!10, draw=black!50] (axis cs:2.75, 71.8) rectangle (axis cs:3.25, 94.6);
    \draw[thick, draw=black!50] (axis cs:2.75, 90.0) -- (axis cs:3.25, 90.0);

    \draw[thick, fill=blue!10, draw=black!50] (axis cs:3.75, 78.4) rectangle (axis cs:4.25, 94.1);
    \draw[thick, draw=black!50] (axis cs:3.75, 90.0) -- (axis cs:4.25, 90.0);

    \addplot[only marks, mark=*, mark size=1.5pt, color=green!70!black] coordinates {(1,94.8) (2,93.9) (3,94.6) (4,94.1)}; 
    \addplot[only marks, mark=diamond*, mark size=2pt, color=green!70!black] coordinates {(1,92.0) (2,91.7) (3,92.3) (4,91.9)}; 
    \addplot[only marks, mark=triangle*, mark size=2pt, color=green!70!black] coordinates {(1,88.3) (2,87.9) (3,88.9) (4,89.2)}; 

    \addplot[only marks, mark=square*, mark size=1.5pt, color=blue!70!black] coordinates {(1,90.4) (2,90.1) (3,91.4) (4,90.8)}; 
    \addplot[only marks, mark=otimes, mark size=2pt, color=blue!70!black] coordinates {(1,82.1) (2,82.9) (3,84.2) (4,83.5)}; 
    \addplot[only marks, mark=square, mark size=2pt, color=blue!70!black] coordinates {(1,78.3) (2,76.4) (3,71.8) (4,78.4)}; 

    \addplot[only marks, mark=star, mark size=2pt, color=red!70!black] coordinates {(1,92.3) (2,92.9) (3,92.9) (4,92.5)}; 
    \addplot[only marks, mark=+, mark size=2.5pt, color=red!70!black] coordinates {(1,92.6) (2,91.4) (3,93.2) (4,93.0)}; 
    \addplot[only marks, mark=x, mark size=2.5pt, color=red!70!black] coordinates {(1,87.1) (2,88.3) (3,82.5) (4,86.2)}; 
    \addplot[only marks, mark=|, mark size=2.5pt, color=red!70!black] coordinates {(1,83.3) (2,79.1) (3,84.3) (4,83.3)}; 
    \end{axis}
    \end{tikzpicture}
    \end{tabular}
    \caption{Beam search}
    \label{subfig:style_sensitivity_temp1}
\end{subfigure}

\vspace{0.5cm} 

\begin{subfigure}{\textwidth}
    \centering
    \setlength{\tabcolsep}{1pt}
    \begin{tabular}{cccc}
    \begin{tikzpicture}[scale=1.0]
    \begin{axis}[
        title={\small\textbf{AIME}},
        xlabel={\small Style},
        ylabel={\small Accuracy (\%)},
        xmin=0.5, xmax=4.5,
        ymin=0, ymax=35,
        xtick={1,2,3,4},
        xticklabels={\small S1, \small S2, \small S3, \small S4},
        ytick={0,10,20,30},
        yticklabels={\small 0, \small 10, \small 20, \small 30},
        width=4cm,
        height=4.5cm,
        axis line style={draw=black!50},
        tick style={draw=black!50},
        grid=none,
    ]
    \draw[thick, fill=blue!10, draw=black!50] (axis cs:0.75, 3.3) rectangle (axis cs:1.25, 26.7);
    \draw[thick, draw=black!50] (axis cs:0.75, 15.0) -- (axis cs:1.25, 15.0);

    \draw[thick, fill=blue!10, draw=black!50] (axis cs:1.75, 3.3) rectangle (axis cs:2.25, 30.0);
    \draw[thick, draw=black!50] (axis cs:1.75, 13.3) -- (axis cs:2.25, 13.3);

    \draw[thick, fill=blue!10, draw=black!50] (axis cs:2.75, 3.3) rectangle (axis cs:3.25, 30.0);
    \draw[thick, draw=black!50] (axis cs:2.75, 11.65) -- (axis cs:3.25, 11.65);

    \draw[thick, fill=blue!10, draw=black!50] (axis cs:3.75, 3.3) rectangle (axis cs:4.25, 26.7);
    \draw[thick, draw=black!50] (axis cs:3.75, 13.35) -- (axis cs:4.25, 13.35);

    \addplot[only marks, mark=*, mark size=1.5pt, color=green!70!black] coordinates {(1,26.7) (2,30.0) (3,30.0) (4,23.3)}; 
    \addplot[only marks, mark=diamond*, mark size=2pt, color=green!70!black] coordinates {(1,3.3) (2,3.3) (3,3.3) (4,3.3)}; 
    \addplot[only marks, mark=triangle*, mark size=2pt, color=green!70!black] coordinates {(1,13.3) (2,13.3) (3,6.7) (4,10.0)}; 

    \addplot[only marks, mark=square*, mark size=1.5pt, color=blue!70!black] coordinates {(1,26.7) (2,23.3) (3,30.0) (4,26.7)}; 
    \addplot[only marks, mark=otimes, mark size=2pt, color=blue!70!black] coordinates {(1,6.7) (2,3.3) (3,10.0) (4,10.0)}; 
    \addplot[only marks, mark=square, mark size=2pt, color=blue!70!black] coordinates {(1,10.0) (2,13.3) (3,13.3) (4,10.0)}; 

    \addplot[only marks, mark=star, mark size=2pt, color=red!70!black] coordinates {(1,20.0) (2,26.7) (3,20.0) (4,20.0)}; 
    \addplot[only marks, mark=+, mark size=2.5pt, color=red!70!black] coordinates {(1,16.7) (2,10.0) (3,20.0) (4,20.0)}; 
    \addplot[only marks, mark=x, mark size=2.5pt, color=red!70!black] coordinates {(1,20.0) (2,16.7) (3,10.0) (4,16.7)}; 
    \addplot[only marks, mark=|, mark size=2.5pt, color=red!70!black] coordinates {(1,6.7) (2,13.3) (3,10.0) (4,3.3)}; 
    \end{axis}
    \end{tikzpicture}
    &
    \begin{tikzpicture}[scale=1.0]
    \begin{axis}[
        title={\small\textbf{MATH-500}},
        xlabel={\small Style},
        ylabel={},
        xmin=0.5, xmax=4.5,
        ymin=38, ymax=80,
        xtick={1,2,3,4},
        xticklabels={\small S1, \small S2, \small S3, \small S4},
        ytick={40,50,60,70,80},
        yticklabels={\small 40, \small 50, \small 60, \small 70, \small 80},
        width=4cm,
        height=4.5cm,
        axis line style={draw=black!50},
        tick style={draw=black!50},
        grid=none,
    ]
    \draw[thick, fill=blue!10, draw=black!50] (axis cs:0.75, 41.6) rectangle (axis cs:1.25, 76.2);
    \draw[thick, draw=black!50] (axis cs:0.75, 65.3) -- (axis cs:1.25, 65.3);

    \draw[thick, fill=blue!10, draw=black!50] (axis cs:1.75, 42.6) rectangle (axis cs:2.25, 76.4);
    \draw[thick, draw=black!50] (axis cs:1.75, 64.4) -- (axis cs:2.25, 64.4);

    \draw[thick, fill=blue!10, draw=black!50] (axis cs:2.75, 42.4) rectangle (axis cs:3.25, 76.8);
    \draw[thick, draw=black!50] (axis cs:2.75, 63.2) -- (axis cs:3.25, 63.2);

    \draw[thick, fill=blue!10, draw=black!50] (axis cs:3.75, 44.8) rectangle (axis cs:4.25, 78.0);
    \draw[thick, draw=black!50] (axis cs:3.75, 64.6) -- (axis cs:4.25, 64.6);

    \addplot[only marks, mark=*, mark size=1.5pt, color=green!70!black] coordinates {(1,76.2) (2,76.4) (3,76.8) (4,78.0)}; 
    \addplot[only marks, mark=diamond*, mark size=2pt, color=green!70!black] coordinates {(1,52.4) (2,52.6) (3,54.6) (4,50.8)}; 
    \addplot[only marks, mark=triangle*, mark size=2pt, color=green!70!black] coordinates {(1,66.6) (2,65.8) (3,64.6) (4,64.6)}; 

    \addplot[only marks, mark=square*, mark size=1.5pt, color=blue!70!black] coordinates {(1,64.4) (2,63.0) (3,61.8) (4,65.0)}; 
    \addplot[only marks, mark=otimes, mark size=2pt, color=blue!70!black] coordinates {(1,43.2) (2,45.0) (3,43.6) (4,44.8)}; 
    \addplot[only marks, mark=square, mark size=2pt, color=blue!70!black] coordinates {(1,41.6) (2,42.6) (3,42.4) (4,45.4)}; 

    \addplot[only marks, mark=star, mark size=2pt, color=red!70!black] coordinates {(1,72.2) (2,70.8) (3,71.8) (4,69.6)}; 
    \addplot[only marks, mark=+, mark size=2.5pt, color=red!70!black] coordinates {(1,71.6) (2,70.6) (3,72.2) (4,69.0)}; 
    \addplot[only marks, mark=x, mark size=2.5pt, color=red!70!black] coordinates {(1,66.2) (2,66.6) (3,65.8) (4,64.6)}; 
    \addplot[only marks, mark=|, mark size=2.5pt, color=red!70!black] coordinates {(1,57.4) (2,55.2) (3,58.2) (4,57.2)}; 
    \end{axis}
    \end{tikzpicture}
    &
    \begin{tikzpicture}[scale=1.0]
    \begin{axis}[
        title={\small\textbf{GPQA-Diamond}},
        xlabel={\small Style},
        ylabel={},
        xmin=0.5, xmax=4.5,
        ymin=1.5, ymax=9,
        xtick={1,2,3,4},
        xticklabels={\small S1, \small S2, \small S3, \small S4},
        ytick={2,4,6,8},
        yticklabels={\small 2, \small 4, \small 6, \small 8},
        width=4cm,
        height=4.5cm,
        axis line style={draw=black!50},
        tick style={draw=black!50},
        grid=none,
    ]
    \draw[thick, fill=blue!10, draw=black!50] (axis cs:0.75, 1.5) rectangle (axis cs:1.25, 7.1);
    \draw[thick, draw=black!50] (axis cs:0.75, 4.3) -- (axis cs:1.25, 4.3);

    \draw[thick, fill=blue!10, draw=black!50] (axis cs:1.75, 1.5) rectangle (axis cs:2.25, 8.1);
    \draw[thick, draw=black!50] (axis cs:1.75, 4.55) -- (axis cs:2.25, 4.55);

    \draw[thick, fill=blue!10, draw=black!50] (axis cs:2.75, 2.5) rectangle (axis cs:3.25, 7.1);
    \draw[thick, draw=black!50] (axis cs:2.75, 3.25) -- (axis cs:3.25, 3.25);

    \draw[thick, fill=blue!10, draw=black!50] (axis cs:3.75, 1.5) rectangle (axis cs:4.25, 7.1);
    \draw[thick, draw=black!50] (axis cs:3.75, 4.55) -- (axis cs:4.25, 4.55);

    \addplot[only marks, mark=*, mark size=1.5pt, color=green!70!black] coordinates {(1,6.1) (2,5.6) (3,5.1) (4,6.6)}; 
    \addplot[only marks, mark=diamond*, mark size=2pt, color=green!70!black] coordinates {(1,6.1) (2,6.6) (3,3.0) (4,5.1)}; 
    \addplot[only marks, mark=triangle*, mark size=2pt, color=green!70!black] coordinates {(1,3.5) (2,2.5) (3,3.0) (4,2.0)}; 

    \addplot[only marks, mark=square*, mark size=1.5pt, color=blue!70!black] coordinates {(1,7.1) (2,8.1) (3,7.1) (4,7.1)}; 
    \addplot[only marks, mark=otimes, mark size=2pt, color=blue!70!black] coordinates {(1,3.5) (2,5.1) (3,3.5) (4,3.0)}; 
    \addplot[only marks, mark=square, mark size=2pt, color=blue!70!black] coordinates {(1,3.0) (2,4.0) (3,2.5) (4,5.1)}; 

    \addplot[only marks, mark=star, mark size=2pt, color=red!70!black] coordinates {(1,5.6) (2,3.5) (3,6.1) (4,5.1)}; 
    \addplot[only marks, mark=+, mark size=2.5pt, color=red!70!black] coordinates {(1,5.1) (2,6.6) (3,7.1) (4,4.0)}; 
    \addplot[only marks, mark=x, mark size=2.5pt, color=red!70!black] coordinates {(1,2.0) (2,2.0) (3,3.0) (4,1.5)}; 
    \addplot[only marks, mark=|, mark size=2.5pt, color=red!70!black] coordinates {(1,1.5) (2,1.5) (3,3.0) (4,3.0)}; 
    \end{axis}
    \end{tikzpicture}
    &
    \begin{tikzpicture}[scale=1.0]
    \begin{axis}[
        title={\small\textbf{GSM8K}},
        xlabel={\small Style},
        ylabel={},
        xmin=0.5, xmax=4.5,
        ymin=70, ymax=96,
        xtick={1,2,3,4},
        xticklabels={\small S1, \small S2, \small S3, \small S4},
        ytick={70,80,90},
        yticklabels={\small 70, \small 80, \small 90},
        width=4cm,
        height=4.5cm,
        axis line style={draw=black!50},
        tick style={draw=black!50},
        grid=none,
    ]
    \draw[thick, fill=blue!10, draw=black!50] (axis cs:0.75, 79.8) rectangle (axis cs:1.25, 94.5);
    \draw[thick, draw=black!50] (axis cs:0.75, 90.25) -- (axis cs:1.25, 90.25);

    \draw[thick, fill=blue!10, draw=black!50] (axis cs:1.75, 79.1) rectangle (axis cs:2.25, 94.2);
    \draw[thick, draw=black!50] (axis cs:1.75, 89.4) -- (axis cs:2.25, 89.4);

    \draw[thick, fill=blue!10, draw=black!50] (axis cs:2.75, 78.0) rectangle (axis cs:3.25, 94.2);
    \draw[thick, draw=black!50] (axis cs:2.75, 89.85) -- (axis cs:3.25, 89.85);

    \draw[thick, fill=blue!10, draw=black!50] (axis cs:3.75, 79.8) rectangle (axis cs:4.25, 94.6);
    \draw[thick, draw=black!50] (axis cs:3.75, 90.15) -- (axis cs:4.25, 90.15);

    \addplot[only marks, mark=*, mark size=1.5pt, color=green!70!black] coordinates {(1,94.5) (2,94.2) (3,94.2) (4,94.6)}; 
    \addplot[only marks, mark=diamond*, mark size=2pt, color=green!70!black] coordinates {(1,90.0) (2,89.9) (3,90.3) (4,91.0)}; 
    \addplot[only marks, mark=triangle*, mark size=2pt, color=green!70!black] coordinates {(1,87.1) (2,84.8) (3,84.8) (4,86.1)}; 

    \addplot[only marks, mark=square*, mark size=1.5pt, color=blue!70!black] coordinates {(1,93.7) (2,93.3) (3,93.2) (4,93.3)}; 
    \addplot[only marks, mark=otimes, mark size=2pt, color=blue!70!black] coordinates {(1,84.7) (2,85.2) (3,84.9) (4,84.9)}; 
    \addplot[only marks, mark=square, mark size=2pt, color=blue!70!black] coordinates {(1,79.8) (2,79.1) (3,78.0) (4,79.8)}; 

    \addplot[only marks, mark=star, mark size=2pt, color=red!70!black] coordinates {(1,92.8) (2,92.3) (3,93.0) (4,93.3)}; 
    \addplot[only marks, mark=+, mark size=2.5pt, color=red!70!black] coordinates {(1,92.8) (2,92.8) (3,92.3) (4,92.7)}; 
    \addplot[only marks, mark=x, mark size=2.5pt, color=red!70!black] coordinates {(1,90.5) (2,88.9) (3,89.4) (4,89.3)}; 
    \addplot[only marks, mark=|, mark size=2.5pt, color=red!70!black] coordinates {(1,84.7) (2,84.8) (3,84.2) (4,84.6)}; 
    \end{axis}
    \end{tikzpicture}
    \end{tabular}
    \caption{Greedy search}
    \label{subfig:style_sensitivity_temp0}
\end{subfigure}

\vspace{0.2cm} 
\small
\centering
\begin{tabular}{cccccc}
\multicolumn{2}{c}{\textbf{Gemma 3}} & \multicolumn{2}{c}{\textbf{LLaMA 3+}} & \multicolumn{2}{c}{\textbf{Qwen 2.5}} \\
\textcolor{green!70!black}{$\bullet$} 27B & 
\textcolor{green!70!black}{$\blacklozenge$} 12B & 
\textcolor{blue!70!black}{$\blacksquare$} 70B & 
\textcolor{blue!70!black}{$\otimes$} 8B & 
\textcolor{red!70!black}{$\star$} 72B & 
\textcolor{red!70!black}{$+$} 32B \\
\textcolor{green!70!black}{$\blacktriangle$} 4B & 
 & 
\textcolor{blue!70!black}{$\square$} 3B & 
 & 
\textcolor{red!70!black}{$\times$} 7B & 
\textcolor{red!70!black}{$|$} 3B
\end{tabular}

\caption{Instruction style sensitivity: Accuracy ranges (S1-S4) by model family and size for Beam Search (a, T=1.0) and Greedy Search (b, T=0.0), showing performance impact of instruction style.}
\label{fig:combined_style_sensitivity}
\end{figure*}
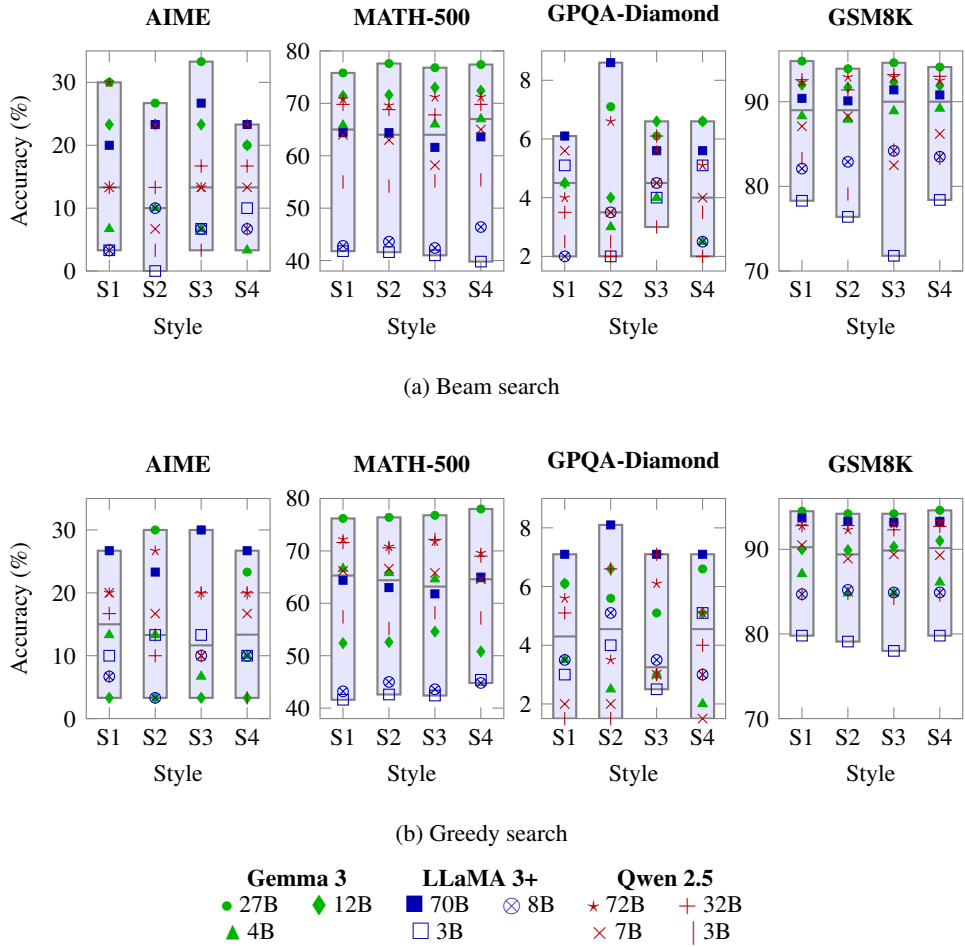

\subsection{Experimental Settings} 
We evaluated models using two decoding strategies: beam search and greedy search, both with \texttt{max\_new\_tokens = 2048}. For beam search, we used \texttt{temperature = 1.0}, \texttt{top-k = 50}, and \texttt{top-p = 0.9}. Greedy search involved selecting the highest probability token at each step. We included both decoding strategies to comprehensively examine model behavior across a broad range of generation settings.

\subsection{Result}
\label{sec:experiment_results}

Our experiments demonstrate performance variations across instruction styles and decoding strategies (Figure \ref{fig:combined_style_sensitivity}). Models exhibit accuracy fluctuations when identical problems are framed using different instruction styles (S1: Declarative, S2: Interrogative, S3: Exclamative, S4: Imperative, as per Figure \ref{fig:instruction_styles}). Detailed Style Sensitivity Index scores across all benchmarks are provided in Table~\ref{tab:all_benchmarks_sensitivity_final}, with comprehensive benchmark-specific analyses available in Appendix \ref{sec:style_sensitivity_analysis}.

Under beam search (Figure \ref{fig:combined_style_sensitivity}a), accuracy variations were pronounced. For instance, on AIME, Gemma 3-27B and Qwen 2.5-72B exhibited gaps of 13.3\% and 16.7\%, respectively, while LLaMA 3-70B showed a 3.0\% gap on GPQA-Diamond. Under greedy search (Figure~\ref{fig:combined_style_sensitivity}b), these variations remained with reduced magnitude—for example, LLaMA 3-70B and Qwen 2.5-72B each shifted by 6.7\% on AIME. This confirms that even deterministic decoding remains susceptible to instruction style.

\begin{figure*}[!t]
    \centering
\begin{tikzpicture}
    \begin{scope}[xshift=-5.5cm]
        \node[align=center] at (0,2.5) {\textbf{Structurality}};
        \node[align=center, font=\small, text width=3.5cm] at (0,1.5) {\textit{"To what extent do responses maintain consistent syntactic patterns?"}};
        
        \draw[rounded corners] (-1.5,0) rectangle (1.5,1);
        \draw[rounded corners] (-1.5,-1.5) rectangle (1.5,-0.5);
        
        \draw[blue!60, thin] (-1.3,0.85) -- (-1.3,0.65);
        \draw[blue!60, thin] (-1.3,0.65) -- (-1.5,0.45);
        \draw[blue!60, thin] (-1.3,0.65) -- (-1.1,0.45);
        \node[font=\tiny] at (-1.3,0.85) {S};
        \node[font=\tiny] at (-1.5,0.45) {NP};
        \node[font=\tiny] at (-1.1,0.45) {VP};
        \draw[blue!60, thin] (-1.5,0.45) -- (-1.5,0.25);
        \draw[blue!60, thin] (-1.1,0.45) -- (-1.1,0.25);
        
        \draw[blue!60, thin] (0,0.85) -- (0,0.65);
        \draw[blue!60, thin] (0,0.65) -- (-0.3,0.45);
        \draw[blue!60, thin] (0,0.65) -- (0.3,0.45);
        \node[font=\tiny] at (0,0.85) {S};
        \node[font=\tiny] at (-0.3,0.45) {NP};
        \node[font=\tiny] at (0.3,0.45) {VP};
        \draw[blue!60, thin] (-0.3,0.45) -- (-0.3,0.25);
        \draw[blue!60, thin] (0.3,0.45) -- (0.1,0.25);
        \draw[blue!60, thin] (0.3,0.45) -- (0.5,0.25);
        \node[font=\tiny] at (0.5,0.25) {PP};
        
        \draw[blue!60, thin] (1.3,0.85) -- (1.3,0.65);
        \draw[blue!60, thin] (1.3,0.65) -- (1.0,0.45);
        \draw[blue!60, thin] (1.3,0.65) -- (1.45,0.45);
        \node[font=\tiny] at (1.3,0.85) {S};
        \node[font=\tiny] at (1.0,0.45) {ADV};
        \node[font=\tiny] at (1.45,0.45) {VP};
        \draw[blue!60, thin] (1.45,0.45) -- (1.45,0.25);
        
        \draw[blue!60, thin] (-1.3,-0.65) -- (-1.3,-0.85);
        \draw[blue!60, thin] (-1.3,-0.85) -- (-1.5,-1.05);
        \draw[blue!60, thin] (-1.3,-0.85) -- (-1.1,-1.05);
        \node[font=\tiny] at (-1.3,-0.65) {S};
        \node[font=\tiny] at (-1.5,-1.05) {NP};
        \node[font=\tiny] at (-1.1,-1.05) {VP};
        \draw[blue!60, thin] (-1.5,-1.05) -- (-1.5,-1.25);
        \draw[blue!60, thin] (-1.1,-1.05) -- (-1.1,-1.25);
        
        \draw[blue!60, thin] (0,-0.65) -- (0,-0.85);
        \draw[blue!60, thin] (0,-0.85) -- (-0.3,-1.05);
        \draw[blue!60, thin] (0,-0.85) -- (0.3,-1.05);
        \node[font=\tiny] at (0,-0.65) {S};
        \node[font=\tiny] at (-0.3,-1.05) {NP};
        \node[font=\tiny] at (0.3,-1.05) {VP};
        \draw[blue!60, thin] (-0.3,-1.05) -- (-0.3,-1.25);
        \draw[blue!60, thin] (0.3,-1.05) -- (0.1,-1.25);
        \draw[blue!60, thin] (0.3,-1.05) -- (0.5,-1.25);
        \node[font=\tiny] at (0.5,-1.25) {PP};
        
        \draw[blue!60, thin] (1.3,-0.65) -- (1.3,-0.85);
        \draw[blue!60, thin] (1.3,-0.85) -- (1.0,-1.05);
        \draw[blue!60, thin] (1.3,-0.85) -- (1.45,-1.05);
        \node[font=\tiny] at (1.3,-0.65) {S};
        \node[font=\tiny] at (1.0,-1.05) {ADV};
        \node[font=\tiny] at (1.45,-1.05) {VP};
        \draw[blue!60, thin] (1.45,-1.05) -- (1.45,-1.25);
        
        \draw[dashed, blue!40] (-1.3,0.65) -- (-1.3,-0.85);
        \draw[dashed, blue!40] (0,0.65) -- (0,-0.85);
        \draw[dashed, blue!40] (1.3,0.65) -- (1.3,-0.85);
        
        \node[align=center, font=\footnotesize, text width=3cm] at (0,-2.2) {Measures syntactic pattern preservation via dependency parsing};
    \end{scope}
    
    \begin{scope}[xshift=0cm]
        \node[align=center] at (0,2.5) {\textbf{Lexicality}};
        \node[align=center, font=\small, text width=3.5cm] at (0,1.5) {\textit{"How consistently do responses use similar vocabulary and terminology?"}};
        
        \draw[rounded corners] (-1.5,0) rectangle (1.5,1);
        \draw[rounded corners] (-1.5,-1.5) rectangle (1.5,-0.5);
        
        \foreach \i/\x/\y/\c/\xx/\yy in {
            1/-1.3/0.7/red!70/-1.3/-0.7, 
            2/-0.9/0.3/green!70/-1.0/-1.1, 
            3/-0.5/0.8/blue!70/-0.6/-0.9, 
            4/-0.1/0.5/orange!70/0.1/-0.8,
            5/0.3/0.2/purple!70/0.5/-1.2,
            6/0.7/0.7/teal!70/0.9/-0.7,
            7/1.2/0.4/brown!70/1.2/-1.0} {
            \filldraw[\c] (\x,\y) circle (0.08);
            \filldraw[\c] (\xx,\yy) circle (0.08);
            \draw[dashed, \c] (\x,\y) -- (\xx,\yy);
        }
        
        \node[align=center, font=\footnotesize, text width=3cm] at (0,-2.2) {Combines TF-IDF vector similarity with sequential matching};
    \end{scope}
    
    \begin{scope}[xshift=5.5cm]
        \node[align=center] at (0,2.5) {\textbf{Coherence}};
        \node[align=center, font=\small, text width=3.5cm] at (0,1.5) {\textit{"Do responses maintain consistent logical progression and organization?"}};
        
        \draw[rounded corners] (-1.5,0) rectangle (1.5,1);
        \draw[rounded corners] (-1.5,-1.5) rectangle (1.5,-0.5);
        
        \def\blockwidth{0.5}
        
        \draw[thick, fill=blue!15] (-1.4,0.85) rectangle (-0.9,0.25);
        \draw[thick, fill=green!15] (-0.8,0.85) rectangle (-0.3,0.25);
        \draw[thick, fill=red!15] (-0.2,0.85) rectangle (0.3,0.25);
        \draw[thick, fill=yellow!15] (0.4,0.85) rectangle (0.9,0.25);
        \draw[thick, fill=purple!15] (1.0,0.85) rectangle (1.4,0.25);
        
        \draw[thick, fill=green!20] (-1.4,-0.65) rectangle (-0.9,-1.25);  
        \draw[thick, fill=blue!20] (-0.8,-0.65) rectangle (-0.3,-1.25);   
        \draw[thick, fill=purple!20] (-0.2,-0.65) rectangle (0.3,-1.25);  
        \draw[thick, fill=red!20] (0.4,-0.65) rectangle (0.9,-1.25);      
        \draw[thick, fill=yellow!20] (1.0,-0.65) rectangle (1.4,-1.25);   
        
        \draw[->, thick, dashed] (-1.15,0.25) -- (-0.55,-0.65);  
        \draw[->, thick, dashed] (-0.55,0.25) -- (-1.15,-0.65);  
        \draw[->, thick, dashed] (0.05,0.25) -- (0.65,-0.65);    
        \draw[->, thick, dashed] (0.65,0.25) -- (1.2,-0.65);     
        \draw[->, thick, dashed] (1.2,0.25) -- (0.05,-0.65);     
        
        \node[align=center, font=\footnotesize, text width=3cm] at (0,-2.2) {Assesses logical flow and content organization preservation};
    \end{scope}
\end{tikzpicture}
\caption{\textbf{RCScore}: A multi-dimensional metric for quantifying response consistency across instruction styles.}
    \label{fig:stylemetrics_dimensions}
\end{figure*}
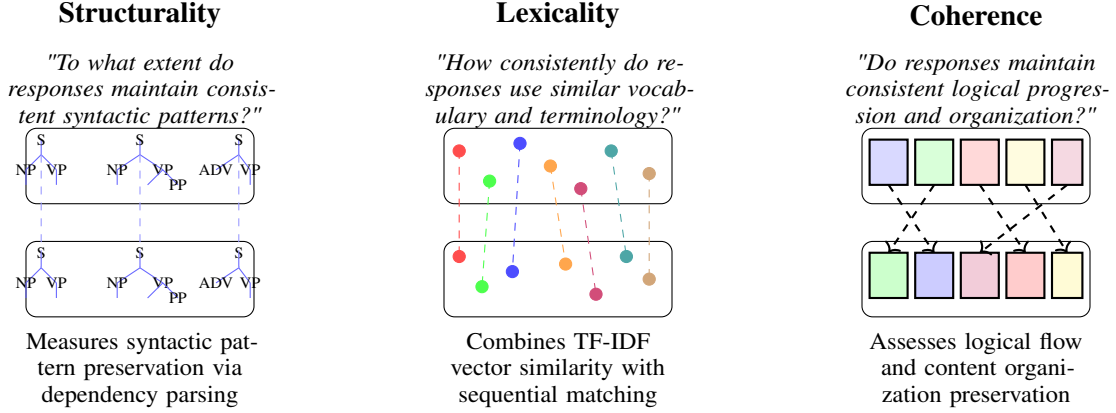

These findings underscore that single-style benchmarks offer an incomplete view of model capabilities. We posit that instruction-induced accuracy fluctuations reflect differences in \textit{response consistency}—the degree to which a model produces structurally, lexically, and logically stable outputs across prompt styles.

\section{RCScore: Quantifying Response Consistency Across Instruction Styles}

The observed accuracy fluctuations, as detailed in Section \ref{sec:experiment_results}, highlight the limitations of relying solely on task performance metrics. To enable a more precise and multi-faceted analysis of how instruction style influences the fundamental qualities of text generated by LLMs, we introduce RCScore, a comprehensive framework that measures response variation across stylistic dimensions. Unlike conventional metrics focused purely on correctness, RCScore captures how different prompt formulations impact model behavior through three complementary dimensions—Structurality, Lexicality, and Coherence—each quantifying a core aspect of stylistic variation (Figure~\ref{fig:stylemetrics_dimensions}). This enables structured evaluation of instruction sensitivity by examining stylistic consistency and assessing fidelity to reference outputs under diverse prompt formulations.

\paragraph{Structurality}
Structurality quantifies the syntactic similarity between text samples, addressing the research question: \textit{"To what extent do responses maintain consistent syntactic patterns and grammatical structures?"} This dimension is formalized as:
\begin{align}
S(D_a, D_b) &= \frac{1}{|M|}\sum_{(s_i, t_i) \in M} J(P(s_i), P(t_i))
\end{align}
Here, $M$ denotes semantically aligned sentence pairs (identified via BERTScore\footnote{We use the RoBERTa-Large model~\cite{liu2019robertarobustlyoptimizedbert} for our BERTScore implementation.}), $P(s)$ extracts syntactic patterns of the form $\langle pos_t, dep_r, pos_h \rangle$, capturing part-of-speech tags and dependency relations, and $J$ computes the Jaccard similarity between the resulting pattern sets.

\paragraph{Lexicality}
Lexicality captures the degree of lexical overlap and vocabulary similarity between generated responses, aiming to answer the question: \textit{"How consistently do responses employ similar vocabulary, terminology, and phrasing?"} We define this dimension as follows:
\begin{align}
L(D_a, D_b) = w_{\text{TF}} \cdot S_{\text{TF}} + w_{\text{RL}} \cdot S_{\text{RL}}
\end{align}
Here, $S_{\text{TF}}$ denotes the cosine similarity between TF-IDF vectors, which reflects the global term distribution across documents. In addition, $S_{\text{RL}}$ measures sequential lexical overlap via ROUGE-L, capturing local word order. By combining these complementary perspectives, Lexicality provides a more comprehensive measure of textual similarity. We assign equal weights ($w_{\text{TF}} = w_{\text{RL}} = 0.5$) to balance global and local contributions.

\paragraph{Coherence}
Coherence assesses the logical flow and organizational similarity between model outputs, addressing the question: \textit{"To what extent do responses maintain consistent logical progression and content organization?"} We define the coherence score as:
\begin{align}
C(D_a, D_b) = S(D_a, D_b) \cdot C_w(D_a, D_b)
\end{align}
The primary component $S(D_a, D_b)$ aggregates four distinct aspects of organizational similarity: order correlation, position matching, sequential continuity, and semantic alignment. Each aspect captures a different facet of how well the content structure is preserved between responses. To ensure comparisons remain semantically meaningful, we apply a content-weighted penalty term $C_w(D_a, D_b)$ that down-weights alignments over irrelevant or trivial content.

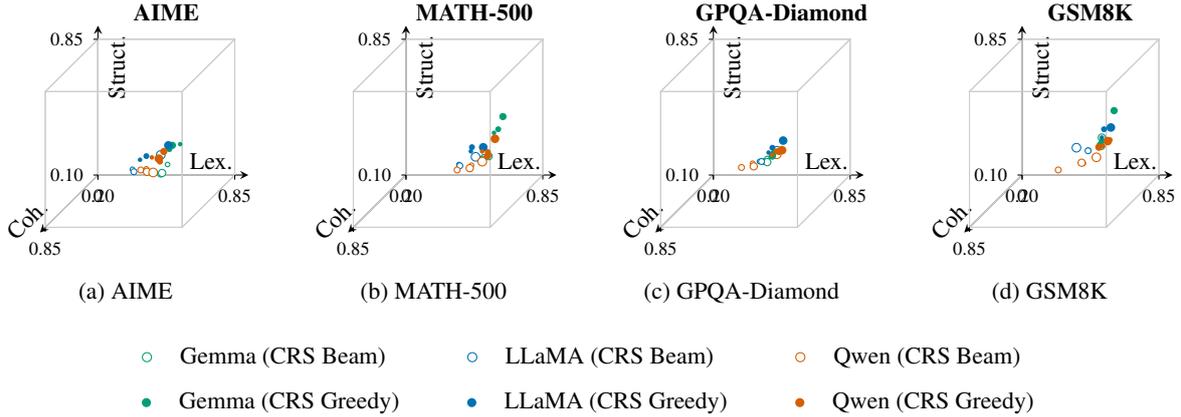
\begin{figure*}[!t]
\centering
\definecolor{gemmablue}{RGB}{0,158,115} 
\definecolor{llamagreen}{RGB}{0,114,178} 
\definecolor{qwenred}{RGB}{213,94,0}    

\begin{subfigure}[b]{0.24\textwidth}
    \centering
    \begin{tikzpicture}[scale=0.6]
        \coordinate (O) at (0,0,0);
        \coordinate (X) at (3.3,0,0);
        \coordinate (Y) at (0,3.3,0);
        \coordinate (Z) at (0,0,3.3);
        
        \draw[-stealth, thin] (O) -- (X) node[anchor=east, xshift=-2pt, yshift=5pt, font=\small] {Lex.};    
        \draw[-stealth, thin] (O) -- (Y) node[anchor=east, rotate=90, yshift=-7pt,font=\small] {Struct.};
        \draw[-stealth, thin] (O) -- (Z) node[anchor=south, rotate=45, font=\small] {Coh.};
        
        \draw[thin, black!20] (0,0,0) -- (3,0,0) -- (3,3,0) -- (0,3,0) -- cycle;
        \draw[thin, black!20] (0,0,0) -- (0,0,3) -- (0,3,3) -- (0,3,0) -- cycle;
        \draw[thin, black!20] (0,0,0) -- (3,0,0) -- (3,0,3) -- (0,0,3) -- cycle;
        \draw[thin, black!20] (3,0,3) -- (3,3,3) -- (0,3,3) -- (0,0,3) -- cycle;
        \draw[thin, black!20] (3,0,0) -- (3,3,0) -- (3,3,3) -- (3,0,3) -- cycle;
        \draw[thin, black!20] (0,3,0) -- (3,3,0) -- (3,3,3) -- (0,3,3) -- cycle;
        
        \foreach \x/\label in {0/0.20, 3/0.85}
            \draw (\x,0,0) -- (\x,-0.1,0) node[anchor=north, font=\scriptsize] {\label};
        \foreach \y/\label in {0/0.10, 3/0.85}
            \draw (0,\y,0) -- (-0.1,\y,0) node[anchor=east, font=\scriptsize] {\label};
        \foreach \z/\label in {0/0, 3/0.85}
            \draw (0,0,\z) -- (0,-0.1,\z) node[anchor=north, font=\scriptsize] {\label};
        
        \draw[gemmablue, fill=white] ( {((0.61 - 0.20) / 0.65) * 3.0}, {((0.25 - 0.10) / 0.75) * 3.0}, { (0.27 / 0.85) * 3.0} ) circle (0.05);  
        \draw[gemmablue, fill=white] ( {((0.58 - 0.20) / 0.65) * 3.0}, {((0.20 - 0.10) / 0.75) * 3.0}, { (0.28 / 0.85) * 3.0} ) circle (0.065); 
        \draw[gemmablue, fill=white] ( {((0.59 - 0.20) / 0.65) * 3.0}, {((0.21 - 0.10) / 0.75) * 3.0}, { (0.29 / 0.85) * 3.0} ) circle (0.08);  
        \draw[llamagreen, fill=white] ( {((0.44 - 0.20) / 0.65) * 3.0}, {((0.22 - 0.10) / 0.75) * 3.0}, { (0.26 / 0.85) * 3.0} ) circle (0.05);  
        \draw[llamagreen, fill=white] ( {((0.45 - 0.20) / 0.65) * 3.0}, {((0.21 - 0.10) / 0.75) * 3.0}, { (0.27 / 0.85) * 3.0} ) circle (0.065); 
        \draw[llamagreen, fill=white] ( {((0.61 - 0.20) / 0.65) * 3.0}, {((0.34 - 0.10) / 0.75) * 3.0}, { (0.38 / 0.85) * 3.0} ) circle (0.095); 
        \draw[qwenred, fill=white] ( {((0.53 - 0.20) / 0.65) * 3.0}, {((0.25 - 0.10) / 0.75) * 3.0}, { (0.34 / 0.85) * 3.0} ) circle (0.05);  
        \draw[qwenred, fill=white] ( {((0.50 - 0.20) / 0.65) * 3.0}, {((0.24 - 0.10) / 0.75) * 3.0}, { (0.33 / 0.85) * 3.0} ) circle (0.065); 
        \draw[qwenred, fill=white] ( {((0.53 - 0.20) / 0.65) * 3.0}, {((0.23 - 0.10) / 0.75) * 3.0}, { (0.33 / 0.85) * 3.0} ) circle (0.08);  
        \draw[qwenred, fill=white] ( {((0.58 - 0.20) / 0.65) * 3.0}, {((0.25 - 0.10) / 0.75) * 3.0}, { (0.40 / 0.85) * 3.0} ) circle (0.095); 

        \fill[gemmablue] ( {((0.72 - 0.20) / 0.65) * 3.0}, {((0.42 - 0.10) / 0.75) * 3.0}, { (0.44 / 0.85) * 3.0} ) circle (0.05);  
        \fill[gemmablue] ( {((0.71 - 0.20) / 0.65) * 3.0}, {((0.44 - 0.10) / 0.75) * 3.0}, { (0.58 / 0.85) * 3.0} ) circle (0.065); 
        \fill[gemmablue] ( {((0.68 - 0.20) / 0.65) * 3.0}, {((0.41 - 0.10) / 0.75) * 3.0}, { (0.43 / 0.85) * 3.0} ) circle (0.08);  
        \fill[llamagreen] ( {((0.50 - 0.20) / 0.65) * 3.0}, {((0.30 - 0.10) / 0.75) * 3.0}, { (0.34 / 0.85) * 3.0} ) circle (0.05);  
        \fill[llamagreen] ( {((0.53 - 0.20) / 0.65) * 3.0}, {((0.32 - 0.10) / 0.75) * 3.0}, { (0.34 / 0.85) * 3.0} ) circle (0.065); 
        \fill[llamagreen] ( {((0.67 - 0.20) / 0.65) * 3.0}, {((0.42 - 0.10) / 0.75) * 3.0}, { (0.46 / 0.85) * 3.0} ) circle (0.095); 
        \fill[qwenred] ( {((0.58 - 0.20) / 0.65) * 3.0}, {((0.34 - 0.10) / 0.75) * 3.0}, { (0.42 / 0.85) * 3.0} ) circle (0.05);  
        \fill[qwenred] ( {((0.63 - 0.20) / 0.65) * 3.0}, {((0.33 - 0.10) / 0.75) * 3.0}, { (0.46 / 0.85) * 3.0} ) circle (0.065); 
        \fill[qwenred] ( {((0.66 - 0.20) / 0.65) * 3.0}, {((0.40 - 0.10) / 0.75) * 3.0}, { (0.50 / 0.85) * 3.0} ) circle (0.08);  
        \fill[qwenred] ( {((0.61 - 0.20) / 0.65) * 3.0}, {((0.33 - 0.10) / 0.75) * 3.0}, { (0.41 / 0.85) * 3.0} ) circle (0.095); 
        
        \node[align=center, anchor=north, font=\small] at (1.5,4.0) {\textbf{AIME}};
    \end{tikzpicture}
    \caption{AIME}
\end{subfigure}%
\hfill
\begin{subfigure}[b]{0.24\textwidth}
    \centering
    \begin{tikzpicture}[scale=0.6]
        \coordinate (O) at (0,0,0);
        \coordinate (X) at (3.3,0,0);
        \coordinate (Y) at (0,3.3,0);
        \coordinate (Z) at (0,0,3.3);
        
        \draw[-stealth, thin] (O) -- (X) node[anchor=east, xshift=-2pt, yshift=5pt, font=\small] {Lex.};      
        \draw[-stealth, thin] (O) -- (Y) node[anchor=east, rotate=90, yshift=-7pt,font=\small] {Struct.};
        \draw[-stealth, thin] (O) -- (Z) node[anchor=south, rotate=45, font=\small] {Coh.};
        
        \draw[thin, black!20] (0,0,0) -- (3,0,0) -- (3,3,0) -- (0,3,0) -- cycle;
        \draw[thin, black!20] (0,0,0) -- (0,0,3) -- (0,3,3) -- (0,3,0) -- cycle;
        \draw[thin, black!20] (0,0,0) -- (3,0,0) -- (3,0,3) -- (0,0,3) -- cycle;
        \draw[thin, black!20] (3,0,3) -- (3,3,3) -- (0,3,3) -- (0,0,3) -- cycle;
        \draw[thin, black!20] (3,0,0) -- (3,3,0) -- (3,3,3) -- (3,0,3) -- cycle;
        \draw[thin, black!20] (0,3,0) -- (3,3,0) -- (3,3,3) -- (0,3,3) -- cycle;
        
        \foreach \x/\label in {0/0.20, 3/0.85}
            \draw (\x,0,0) -- (\x,-0.1,0) node[anchor=north, font=\scriptsize] {\label};
        \foreach \y/\label in {0/0.10, 3/0.85}
            \draw (0,\y,0) -- (-0.1,\y,0) node[anchor=east, font=\scriptsize] {\label};
        \foreach \z/\label in {0/0, 3/0.85}
            \draw (0,0,\z) -- (0,-0.1,\z) node[anchor=north, font=\scriptsize] {\label};

        \draw[gemmablue, fill=white] ( {((0.68 - 0.20) / 0.65) * 3.0}, {((0.33 - 0.10) / 0.75) * 3.0}, { (0.38 / 0.85) * 3.0} ) circle (0.05);  
        \draw[gemmablue, fill=white] ( {((0.68 - 0.20) / 0.65) * 3.0}, {((0.32 - 0.10) / 0.75) * 3.0}, { (0.39 / 0.85) * 3.0} ) circle (0.065); 
        \draw[gemmablue, fill=white] ( {((0.71 - 0.20) / 0.65) * 3.0}, {((0.34 - 0.10) / 0.75) * 3.0}, { (0.40 / 0.85) * 3.0} ) circle (0.08);  
        \draw[llamagreen, fill=white] ( {((0.54 - 0.20) / 0.65) * 3.0}, {((0.26 - 0.10) / 0.75) * 3.0}, { (0.30 / 0.85) * 3.0} ) circle (0.05);  
        \draw[llamagreen, fill=white] ( {((0.55 - 0.20) / 0.65) * 3.0}, {((0.26 - 0.10) / 0.75) * 3.0}, { (0.32 / 0.85) * 3.0} ) circle (0.065); 
        \draw[llamagreen, fill=white] ( {((0.66 - 0.20) / 0.65) * 3.0}, {((0.35 - 0.10) / 0.75) * 3.0}, { (0.44 / 0.85) * 3.0} ) circle (0.095); 
        \draw[qwenred, fill=white] ( {((0.62 - 0.20) / 0.65) * 3.0}, {((0.28 - 0.10) / 0.75) * 3.0}, { (0.37 / 0.85) * 3.0} ) circle (0.05);  
        \draw[qwenred, fill=white] ( {((0.54 - 0.20) / 0.65) * 3.0}, {((0.24 - 0.10) / 0.75) * 3.0}, { (0.33 / 0.85) * 3.0} ) circle (0.065); 
        \draw[qwenred, fill=white] ( {((0.60 - 0.20) / 0.65) * 3.0}, {((0.25 - 0.10) / 0.75) * 3.0}, { (0.33 / 0.85) * 3.0} ) circle (0.08);  
        \draw[qwenred, fill=white] ( {((0.68 - 0.20) / 0.65) * 3.0}, {((0.31 - 0.10) / 0.75) * 3.0}, { (0.40 / 0.85) * 3.0} ) circle (0.095); 

        \fill[gemmablue] ( {((0.78 - 0.20) / 0.65) * 3.0}, {((0.52 - 0.10) / 0.75) * 3.0}, { (0.55 / 0.85) * 3.0} ) circle (0.05);  
        \fill[gemmablue] ( {((0.80 - 0.20) / 0.65) * 3.0}, {((0.54 - 0.10) / 0.75) * 3.0}, { (0.55 / 0.85) * 3.0} ) circle (0.065); 
        \fill[gemmablue] ( {((0.84 - 0.20) / 0.65) * 3.0}, {((0.63 - 0.10) / 0.75) * 3.0}, { (0.61 / 0.85) * 3.0} ) circle (0.08);  
        \fill[llamagreen] ( {((0.62 - 0.20) / 0.65) * 3.0}, {((0.36 - 0.10) / 0.75) * 3.0}, { (0.38 / 0.85) * 3.0} ) circle (0.05);  
        \fill[llamagreen] ( {((0.64 - 0.20) / 0.65) * 3.0}, {((0.40 - 0.10) / 0.75) * 3.0}, { (0.43 / 0.85) * 3.0} ) circle (0.065); 
        \fill[llamagreen] ( {((0.72 - 0.20) / 0.65) * 3.0}, {((0.43 - 0.10) / 0.75) * 3.0}, { (0.52 / 0.85) * 3.0} ) circle (0.095); 
        \fill[qwenred] ( {((0.70 - 0.20) / 0.65) * 3.0}, {((0.39 - 0.10) / 0.75) * 3.0}, { (0.47 / 0.85) * 3.0} ) circle (0.05);  
        \fill[qwenred] ( {((0.73 - 0.20) / 0.65) * 3.0}, {((0.39 - 0.10) / 0.75) * 3.0}, { (0.48 / 0.85) * 3.0} ) circle (0.065); 
        \fill[qwenred] ( {((0.73 - 0.20) / 0.65) * 3.0}, {((0.37 - 0.10) / 0.75) * 3.0}, { (0.48 / 0.85) * 3.0} ) circle (0.08);  
        \fill[qwenred] ( {((0.77 - 0.20) / 0.65) * 3.0}, {((0.47 - 0.10) / 0.75) * 3.0}, { (0.50 / 0.85) * 3.0} ) circle (0.095); 
        
        \node[align=center, anchor=north, font=\small] at (1.5,4.0) {\textbf{MATH-500}};
    \end{tikzpicture}
    \caption{MATH-500}
\end{subfigure}%
\hfill
\begin{subfigure}[b]{0.24\textwidth}
    \centering
    \begin{tikzpicture}[scale=0.6]
        \coordinate (O) at (0,0,0);
        \coordinate (X) at (3.3,0,0);
        \coordinate (Y) at (0,3.3,0);
        \coordinate (Z) at (0,0,3.3);
        
        \draw[-stealth, thin] (O) -- (X) node[anchor=east, xshift=-2pt, yshift=5pt, font=\small] {Lex.};       
        \draw[-stealth, thin] (O) -- (Y) node[anchor=east, rotate=90, yshift=-7pt,font=\small] {Struct.};
        \draw[-stealth, thin] (O) -- (Z) node[anchor=south, rotate=45, font=\small] {Coh.};

        \draw[thin, black!20] (0,0,0) -- (3,0,0) -- (3,3,0) -- (0,3,0) -- cycle;
        \draw[thin, black!20] (0,0,0) -- (0,0,3) -- (0,3,3) -- (0,3,0) -- cycle;
        \draw[thin, black!20] (0,0,0) -- (3,0,0) -- (3,0,3) -- (0,0,3) -- cycle;
        \draw[thin, black!20] (3,0,3) -- (3,3,3) -- (0,3,3) -- (0,0,3) -- cycle;
        \draw[thin, black!20] (3,0,0) -- (3,3,0) -- (3,3,3) -- (3,0,3) -- cycle;
        \draw[thin, black!20] (0,3,0) -- (3,3,0) -- (3,3,3) -- (0,3,3) -- cycle;
        
        \foreach \x/\label in {0/0.20, 3/0.85}
            \draw (\x,0,0) -- (\x,-0.1,0) node[anchor=north, font=\scriptsize] {\label};
        \foreach \y/\label in {0/0.10, 3/0.85}
            \draw (0,\y,0) -- (-0.1,\y,0) node[anchor=east, font=\scriptsize] {\label};
        \foreach \z/\label in {0/0, 3/0.85}
            \draw (0,0,\z) -- (0,-0.1,\z) node[anchor=north, font=\scriptsize] {\label};

        \draw[gemmablue, fill=white] ( {((0.54 - 0.20) / 0.65) * 3.0}, {((0.29 - 0.10) / 0.75) * 3.0}, { (0.29 / 0.85) * 3.0} ) circle (0.05);  
        \draw[gemmablue, fill=white] ( {((0.52 - 0.20) / 0.65) * 3.0}, {((0.27 - 0.10) / 0.75) * 3.0}, { (0.28 / 0.85) * 3.0} ) circle (0.065); 
        \draw[gemmablue, fill=white] ( {((0.55 - 0.20) / 0.65) * 3.0}, {((0.28 - 0.10) / 0.75) * 3.0}, { (0.32 / 0.85) * 3.0} ) circle (0.08);  
        \draw[llamagreen, fill=white] ( {((0.51 - 0.20) / 0.65) * 3.0}, {((0.28 - 0.10) / 0.75) * 3.0}, { (0.31 / 0.85) * 3.0} ) circle (0.05);  
        \draw[llamagreen, fill=white] ( {((0.52 - 0.20) / 0.65) * 3.0}, {((0.28 - 0.10) / 0.75) * 3.0}, { (0.31 / 0.85) * 3.0} ) circle (0.065); 
        \draw[llamagreen, fill=white] ( {((0.62 - 0.20) / 0.65) * 3.0}, {((0.37 - 0.10) / 0.75) * 3.0}, { (0.40 / 0.85) * 3.0} ) circle (0.095); 
        \draw[qwenred, fill=white] ( {((0.46 - 0.20) / 0.65) * 3.0}, {((0.25 - 0.10) / 0.75) * 3.0}, { (0.25 / 0.85) * 3.0} ) circle (0.05);  
        \draw[qwenred, fill=white] ( {((0.41 - 0.20) / 0.65) * 3.0}, {((0.23 - 0.10) / 0.75) * 3.0}, { (0.26 / 0.85) * 3.0} ) circle (0.065); 
        \draw[qwenred, fill=white] ( {((0.48 - 0.20) / 0.65) * 3.0}, {((0.25 - 0.10) / 0.75) * 3.0}, { (0.30 / 0.85) * 3.0} ) circle (0.08);  
        \draw[qwenred, fill=white] ( {((0.61 - 0.20) / 0.65) * 3.0}, {((0.34 - 0.10) / 0.75) * 3.0}, { (0.37 / 0.85) * 3.0} ) circle (0.095); 

        \fill[gemmablue] ( {((0.57 - 0.20) / 0.65) * 3.0}, {((0.32 - 0.10) / 0.75) * 3.0}, { (0.31 / 0.85) * 3.0} ) circle (0.05);  
        \fill[gemmablue] ( {((0.58 - 0.20) / 0.65) * 3.0}, {((0.33 - 0.10) / 0.75) * 3.0}, { (0.32 / 0.85) * 3.0} ) circle (0.065); 
        \fill[gemmablue] ( {((0.58 - 0.20) / 0.65) * 3.0}, {((0.32 - 0.10) / 0.75) * 3.0}, { (0.34 / 0.85) * 3.0} ) circle (0.08);  
        \fill[llamagreen] ( {((0.58 - 0.20) / 0.65) * 3.0}, {((0.36 - 0.10) / 0.75) * 3.0}, { (0.40 / 0.85) * 3.0} ) circle (0.05);  
        \fill[llamagreen] ( {((0.61 - 0.20) / 0.65) * 3.0}, {((0.40 - 0.10) / 0.75) * 3.0}, { (0.44 / 0.85) * 3.0} ) circle (0.065); 
        \fill[llamagreen] ( {((0.67 - 0.20) / 0.65) * 3.0}, {((0.45 - 0.10) / 0.75) * 3.0}, { (0.47 / 0.85) * 3.0} ) circle (0.095); 
        \fill[qwenred] ( {((0.59 - 0.20) / 0.65) * 3.0}, {((0.34 - 0.10) / 0.75) * 3.0}, { (0.38 / 0.85) * 3.0} ) circle (0.05);  
        \fill[qwenred] ( {((0.62 - 0.20) / 0.65) * 3.0}, {((0.37 - 0.10) / 0.75) * 3.0}, { (0.41 / 0.85) * 3.0} ) circle (0.065); 
        \fill[qwenred] ( {((0.65 - 0.20) / 0.65) * 3.0}, {((0.39 - 0.10) / 0.75) * 3.0}, { (0.47 / 0.85) * 3.0} ) circle (0.08);  
        \fill[qwenred] ( {((0.65 - 0.20) / 0.65) * 3.0}, {((0.38 - 0.10) / 0.75) * 3.0}, { (0.42 / 0.85) * 3.0} ) circle (0.095); 
        
        \node[align=center, anchor=north, font=\small] at (1.5,4.0) {\textbf{GPQA-Diamond}};
    \end{tikzpicture}
    \caption{GPQA-Diamond}
\end{subfigure}%
\hfill
\begin{subfigure}[b]{0.24\textwidth}
    \centering
    \begin{tikzpicture}[scale=0.6]
        \coordinate (O) at (0,0,0);
        \coordinate (X) at (3.3,0,0);
        \coordinate (Y) at (0,3.3,0);
        \coordinate (Z) at (0,0,3.3);
        
        \draw[-stealth, thin] (O) -- (X) node[anchor=east, xshift=-2pt, yshift=5pt, font=\small] {Lex.};
        \draw[-stealth, thin] (O) -- (Y) node[anchor=east, rotate=90, yshift=-7pt,font=\small] {Struct.};
        \draw[-stealth, thin] (O) -- (Z) node[anchor=south, rotate=45, font=\small] {Coh.};

        \draw[thin, black!20] (0,0,0) -- (3,0,0) -- (3,3,0) -- (0,3,0) -- cycle;
        \draw[thin, black!20] (0,0,0) -- (0,0,3) -- (0,3,3) -- (0,3,0) -- cycle;
        \draw[thin, black!20] (0,0,0) -- (3,0,0) -- (3,0,3) -- (0,0,3) -- cycle;
        \draw[thin, black!20] (3,0,3) -- (3,3,3) -- (0,3,3) -- (0,0,3) -- cycle;
        \draw[thin, black!20] (3,0,0) -- (3,3,0) -- (3,3,3) -- (3,0,3) -- cycle;
        \draw[thin, black!20] (0,3,0) -- (3,3,0) -- (3,3,3) -- (0,3,3) -- cycle;
        
        \foreach \x/\label in {0/0.20, 3/0.85}
            \draw (\x,0,0) -- (\x,-0.1,0) node[anchor=north, font=\scriptsize] {\label};
        \foreach \y/\label in {0/0.10, 3/0.85}
            \draw (0,\y,0) -- (-0.1,\y,0) node[anchor=east, font=\scriptsize] {\label};
        \foreach \z/\label in {0/0, 3/0.85}
            \draw (0,0,\z) -- (0,-0.1,\z) node[anchor=north, font=\scriptsize] {\label};

        \draw[gemmablue, fill=white] ( {((0.70 - 0.20) / 0.65) * 3.0}, {((0.42 - 0.10) / 0.75) * 3.0}, { (0.45 / 0.85) * 3.0} ) circle (0.05);  
        \draw[gemmablue, fill=white] ( {((0.71 - 0.20) / 0.65) * 3.0}, {((0.41 - 0.10) / 0.75) * 3.0}, { (0.45 / 0.85) * 3.0} ) circle (0.065); 
        \draw[gemmablue, fill=white] ( {((0.72 - 0.20) / 0.65) * 3.0}, {((0.47 - 0.10) / 0.75) * 3.0}, { (0.48 / 0.85) * 3.0} ) circle (0.08);  
        \draw[llamagreen, fill=white] ( {((0.63 - 0.20) / 0.65) * 3.0}, {((0.37 - 0.10) / 0.75) * 3.0}, { (0.39 / 0.85) * 3.0} ) circle (0.05);  
        \draw[llamagreen, fill=white] ( {((0.64 - 0.20) / 0.65) * 3.0}, {((0.38 - 0.10) / 0.75) * 3.0}, { (0.43 / 0.85) * 3.0} ) circle (0.065); 
        \draw[llamagreen, fill=white] ( {((0.57 - 0.20) / 0.65) * 3.0}, {((0.38 - 0.10) / 0.75) * 3.0}, { (0.38 / 0.85) * 3.0} ) circle (0.095); 
        \draw[qwenred, fill=white] ( {((0.60 - 0.20) / 0.65) * 3.0}, {((0.29 - 0.10) / 0.75) * 3.0}, { (0.37 / 0.85) * 3.0} ) circle (0.05);  
        \draw[qwenred, fill=white] ( {((0.46 - 0.20) / 0.65) * 3.0}, {((0.23 - 0.10) / 0.75) * 3.0}, { (0.30 / 0.85) * 3.0} ) circle (0.065); 
        \draw[qwenred, fill=white] ( {((0.58 - 0.20) / 0.65) * 3.0}, {((0.28 - 0.10) / 0.75) * 3.0}, { (0.33 / 0.85) * 3.0} ) circle (0.08);  
        \draw[qwenred, fill=white] ( {((0.65 - 0.20) / 0.65) * 3.0}, {((0.31 - 0.10) / 0.75) * 3.0}, { (0.33 / 0.85) * 3.0} ) circle (0.095); 

        \fill[gemmablue] ( {((0.73 - 0.20) / 0.65) * 3.0}, {((0.46 - 0.10) / 0.75) * 3.0}, { (0.54 / 0.85) * 3.0} ) circle (0.05);  
        \fill[gemmablue] ( {((0.73 - 0.20) / 0.65) * 3.0}, {((0.46 - 0.10) / 0.75) * 3.0}, { (0.51 / 0.85) * 3.0} ) circle (0.065); 
        \fill[gemmablue] ( {((0.84 - 0.20) / 0.65) * 3.0}, {((0.69 - 0.10) / 0.75) * 3.0}, { (0.69 / 0.85) * 3.0} ) circle (0.08);  
        \fill[llamagreen] ( {((0.72 - 0.20) / 0.65) * 3.0}, {((0.47 - 0.10) / 0.75) * 3.0}, { (0.48 / 0.85) * 3.0} ) circle (0.05);  
        \fill[llamagreen] ( {((0.76 - 0.20) / 0.65) * 3.0}, {((0.55 - 0.10) / 0.75) * 3.0}, { (0.58 / 0.85) * 3.0} ) circle (0.065); 
        \fill[llamagreen] ( {((0.81 - 0.20) / 0.65) * 3.0}, {((0.58 - 0.10) / 0.75) * 3.0}, { (0.64 / 0.85) * 3.0} ) circle (0.095); 
        \fill[qwenred] ( {((0.75 - 0.20) / 0.65) * 3.0}, {((0.45 - 0.10) / 0.75) * 3.0}, { (0.56 / 0.85) * 3.0} ) circle (0.05);  
        \fill[qwenred] ( {((0.76 - 0.20) / 0.65) * 3.0}, {((0.46 - 0.10) / 0.75) * 3.0}, { (0.49 / 0.85) * 3.0} ) circle (0.065); 
        \fill[qwenred] ( {((0.70 - 0.20) / 0.65) * 3.0}, {((0.41 - 0.10) / 0.75) * 3.0}, { (0.46 / 0.85) * 3.0} ) circle (0.08);  
        \fill[qwenred] ( {((0.73 - 0.20) / 0.65) * 3.0}, {((0.43 - 0.10) / 0.75) * 3.0}, { (0.42 / 0.85) * 3.0} ) circle (0.095); 
        
        \node[align=center, anchor=north, font=\small] at (1.5,4.0) {\textbf{GSM8K}};
    \end{tikzpicture}
    \caption{GSM8K}
\end{subfigure}

\vspace{0.5em} 

\newlength{\legendmarkersep}
\setlength{\legendmarkersep}{1mm} 
\newlength{\legendcolumngap}
\setlength{\legendcolumngap}{0.8cm} 
\newlength{\legendrowgap}
\setlength{\legendrowgap}{0.4cm} 

\begin{center}
\begin{tikzpicture}[scale=1, every node/.style={font=\small}]
    \node[anchor=west] (gemma_crs_beam_marker) at (0,0) {\tikz\draw[gemmablue,fill=white] (0,0) circle (0.06cm);};
    \node[anchor=west] (gemma_crs_beam_text) at ([xshift=\legendmarkersep]gemma_crs_beam_marker.east) {Gemma (CRS Beam)};
    
    \node[anchor=west] (llama_crs_beam_marker) at ([xshift=\legendcolumngap]gemma_crs_beam_text.east) {\tikz\draw[llamagreen,fill=white] (0,0) circle (0.06cm);};
    \node[anchor=west] (llama_crs_beam_text) at ([xshift=\legendmarkersep]llama_crs_beam_marker.east) {LLaMA (CRS Beam)};

    \node[anchor=west] (qwen_crs_beam_marker) at ([xshift=\legendcolumngap]llama_crs_beam_text.east) {\tikz\draw[qwenred,fill=white] (0,0) circle (0.06cm);};
    \node[anchor=west] (qwen_crs_beam_text) at ([xshift=\legendmarkersep]qwen_crs_beam_marker.east) {Qwen (CRS Beam)};

    \node[anchor=west] (gemma_crs_greedy_marker) at ([yshift=-\legendrowgap]gemma_crs_beam_marker.south west) {\tikz\fill[gemmablue] (0,0) circle (0.06cm);};
    \node[anchor=west] (gemma_crs_greedy_text) at ([xshift=\legendmarkersep]gemma_crs_greedy_marker.east) {Gemma (CRS Greedy)};
    
    \node[anchor=west] (llama_crs_greedy_marker) at ([yshift=-\legendrowgap]llama_crs_beam_marker.south west) {\tikz\fill[llamagreen] (0,0) circle (0.06cm);};
    \node[anchor=west] (llama_crs_greedy_text) at ([xshift=\legendmarkersep]llama_crs_greedy_marker.east) {LLaMA (CRS Greedy)};

    \node[anchor=west] (qwen_crs_greedy_marker) at ([yshift=-\legendrowgap]qwen_crs_beam_marker.south west) {\tikz\fill[qwenred] (0,0) circle (0.06cm);};
    \node[anchor=west] (qwen_crs_greedy_text) at ([xshift=\legendmarkersep]qwen_crs_greedy_marker.east) {Qwen (CRS Greedy)};
\end{tikzpicture}
\end{center}

\caption{CRS values (RCScore dimensions: Lexicality, Structurality, Coherence) per benchmark. Beam Search (hollow circles) vs. Greedy Search (filled circles). Marker size reflects model parameter count.}
\label{fig:stylemetrics_all_benchmarks}
\end{figure*}

\begin{figure}[!ht]
    \centering
    \begin{tikzpicture}[scale=0.75]
        \node[align=center] at (0,2.3) {\textbf{RCScore Applied Through the CRS Way}};
        
        \node[draw, rounded corners, fill=blue!10, text=blue!70!black, minimum width=1.8cm] (s1) at (-3.2,0.8) {S1 {\footnotesize(Declarative)}};
        \node[draw, rounded corners, fill=violet!10, text=violet!80!black, minimum width=1.8cm] (s2) at (3.2,0.8) {S2 {\footnotesize(Interrogative)}};
        \node[draw, rounded corners, fill=green!10, text=green!60!black, minimum width=1.8cm] (s3) at (-3.2,-1.5) {S3 {\footnotesize(Exclamative)}};
        \node[draw, rounded corners, fill=orange!10, text=orange!80!black, minimum width=1.8cm] (s4) at (3.2,-1.5) {S4 {\footnotesize(Imperative)}};
        
        \draw[<->, dashed] (s1.east) to[bend left=10] node[midway, above, font=\scriptsize] {CRS$_{1,2}$} (s2.west);
        \draw[<->, dashed] (s1.south) to node[midway, left, font=\scriptsize] {CRS$_{1,3}$} (s3.north);
        
        \draw[<->, dashed] (s1.south east) to[bend right=15] node[pos=0.1, below right, font=\scriptsize] {CRS$_{1,4}$} (s4.north west);
        \draw[<->, dashed] (s2.south west) to[bend left=15] node[pos=0.5, below left, font=\scriptsize] {CRS$_{2,3}$} (s3.north east);
        
        \draw[<->, dashed] (s2.south) to node[midway, right, font=\scriptsize] {CRS$_{2,4}$} (s4.north);
        \draw[<->, dashed] (s3.east) to[bend right=10] node[midway, below, font=\scriptsize] {CRS$_{3,4}$} (s4.west);
        
        \node[draw, rounded corners, align=left, font=\scriptsize, text width=5cm] at (0,-3.2) {
        \textbf{All comparisons yield 3D vectors:}\\
        $\langle$ \textcolor{blue!70!black}{Structurality}, 
        \textcolor{green!70!black}{Lexicality}, 
        \textcolor{red!70!black}{Coherence} $\rangle$\\
        \textbf{Aggregate metrics:}\\
        CRS: Average cross-style similarity (6 pairs)
        };
    \end{tikzpicture}
    \caption{The Cross-Response Similarity (CRS) way applies RCScore dimensions to measure consistency across all possible combinations of instruction styles (6 pairwise comparisons).}
    \label{fig:style_sensitivity_metrics}
\end{figure}
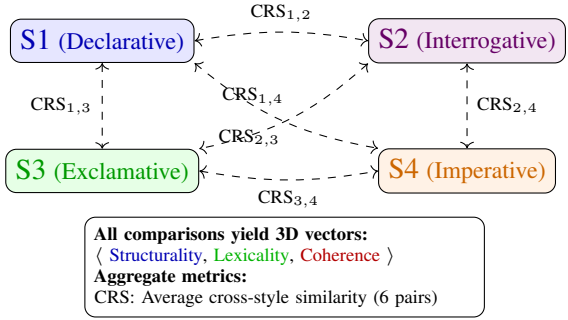

A final RCScore is then computed as a weighted average of these three dimensions, with Structurality, Lexicality, and Coherence each assigned a weight of 0.33.  For practical examples of how these dimensions quantify similarity, refer to Table~\ref{tab:stylemetrics_example_similar_colored} (high similarity) and Table~\ref{tab:style-metrics-different_colored} (low similarity), which demonstrate the metrics' application through concrete cases. A comprehensive mathematical treatment is provided in Appendix~\ref{sec:stylemetrics_computation}.

\subsection{Quantification of Cross-Style Response Consistency using RCScore}
\label{crs_section}

\begin{table*}[t]
\centering

\begin{subtable}{\textwidth}
\centering
\resizebox{\textwidth}{!}{%
\begin{tabular}{@{}l|cccc|cccc|cccc|cccc@{}}
\toprule
\multirow{3}{*}{\textbf{Model}} & \multicolumn{4}{c|}{\textbf{AIME}} & \multicolumn{4}{c|}{\textbf{MATH-500}} & \multicolumn{4}{c|}{\textbf{GPQA-Diamond}} & \multicolumn{4}{c|}{\textbf{GSM8K}} \\
\cmidrule(lr){2-5} \cmidrule(lr){6-9} \cmidrule(lr){10-13} \cmidrule(lr){14-17}
 & \multicolumn{4}{c|}{\textbf{CRS}} & \multicolumn{4}{c|}{\textbf{CRS}} & \multicolumn{4}{c|}{\textbf{CRS}} & \multicolumn{4}{c|}{\textbf{CRS}} \\
\cmidrule(lr){2-5} \cmidrule(lr){6-9} \cmidrule(lr){10-13} \cmidrule(lr){14-17}
 & \textbf{Struct.} & \textbf{Lex.} & \textbf{Coh.} & \textbf{RCScore} & \textbf{Struct.} & \textbf{Lex.} & \textbf{Coh.} & \textbf{RCScore} & \textbf{Struct.} & \textbf{Lex.} & \textbf{Coh.} & \textbf{RCScore} & \textbf{Struct.} & \textbf{Lex.} & \textbf{Coh.} & \textbf{RCScore} \\
\midrule
Gemma 3-4B    & 0.25 & 0.61 & 0.27 & 0.38 & 0.33 & 0.68 & 0.38 & 0.46 & 0.29 & 0.54 & 0.29 & 0.37 & 0.42 & 0.70 & 0.45 & 0.52 \\
Gemma 3-12B   & 0.20 & 0.58 & 0.28 & 0.36 & 0.32 & 0.68 & 0.39 & 0.46 & 0.27 & 0.52 & 0.28 & 0.36 & 0.41 & 0.71 & 0.45 & 0.52 \\
Gemma 3-27B   & 0.21 & 0.59 & 0.29 & 0.37 & 0.34 & 0.71 & 0.40 & 0.48 & 0.28 & 0.55 & 0.32 & 0.38 & 0.47 & 0.72 & 0.48 & 0.56 \\
\hdashline
LLaMA 3.2-3B  & 0.22 & 0.44 & 0.26 & 0.31 & 0.26 & 0.54 & 0.30 & 0.36 & 0.28 & 0.51 & 0.31 & 0.37 & 0.37 & 0.63 & 0.39 & 0.46 \\
LLaMA 3.1-8B  & 0.21 & 0.45 & 0.27 & 0.31 & 0.26 & 0.55 & 0.32 & 0.38 & 0.28 & 0.52 & 0.31 & 0.37 & 0.38 & 0.64 & 0.43 & 0.48 \\
LLaMA 3.3-70B & 0.34 & 0.61 & 0.38 & 0.44 & 0.35 & 0.66 & 0.44 & 0.48 & 0.37 & 0.62 & 0.40 & 0.46 & 0.38 & 0.57 & 0.38 & 0.44 \\
\hdashline
Qwen 2.5-3B   & 0.25 & 0.53 & 0.34 & 0.37 & 0.28 & 0.62 & 0.37 & 0.42 & 0.25 & 0.46 & 0.25 & 0.32 & 0.29 & 0.60 & 0.37 & 0.42 \\
Qwen 2.5-7B   & 0.24 & 0.50 & 0.33 & 0.35 & 0.24 & 0.54 & 0.33 & 0.37 & 0.23 & 0.41 & 0.26 & 0.30 & 0.23 & 0.46 & 0.30 & 0.33 \\
Qwen 2.5-32B  & 0.23 & 0.53 & 0.33 & 0.36 & 0.25 & 0.60 & 0.33 & 0.39 & 0.25 & 0.48 & 0.30 & 0.34 & 0.28 & 0.58 & 0.33 & 0.39 \\
Qwen 2.5-72B  & 0.25 & 0.58 & 0.40 & 0.41 & 0.31 & 0.68 & 0.40 & 0.46 & 0.34 & 0.61 & 0.37 & 0.44 & 0.31 & 0.65 & 0.33 & 0.43 \\
\bottomrule
\end{tabular}%
}
\caption{Beam Search CRS Values (temperature = 1.0).}
\label{tab:stylemetrics_crs_beam_search}
\end{subtable}

\vspace{1em} 

\begin{subtable}{\textwidth}
\centering
\resizebox{\textwidth}{!}{%
\begin{tabular}{@{}l|cccc|cccc|cccc|cccc@{}}
\toprule
\multirow{3}{*}{\textbf{Model}} & \multicolumn{4}{c|}{\textbf{AIME}} & \multicolumn{4}{c|}{\textbf{MATH-500}} & \multicolumn{4}{c|}{\textbf{GPQA-Diamond}} & \multicolumn{4}{c|}{\textbf{GSM8K}} \\
\cmidrule(lr){2-5} \cmidrule(lr){6-9} \cmidrule(lr){10-13} \cmidrule(lr){14-17}
 & \multicolumn{4}{c|}{\textbf{CRS}} & \multicolumn{4}{c|}{\textbf{CRS}} & \multicolumn{4}{c|}{\textbf{CRS}} & \multicolumn{4}{c|}{\textbf{CRS}} \\
\cmidrule(lr){2-5} \cmidrule(lr){6-9} \cmidrule(lr){10-13} \cmidrule(lr){14-17}
 & \textbf{Struct.} & \textbf{Lex.} & \textbf{Coh.} & \textbf{RCScore} & \textbf{Struct.} & \textbf{Lex.} & \textbf{Coh.} & \textbf{RCScore} & \textbf{Struct.} & \textbf{Lex.} & \textbf{Coh.} & \textbf{RCScore} & \textbf{Struct.} & \textbf{Lex.} & \textbf{Coh.} & \textbf{RCScore} \\
\midrule
Gemma 3-4B    & 0.42 & 0.72 & 0.44 & 0.53 & 0.52 & 0.78 & 0.55 & 0.61 & 0.32 & 0.57 & 0.31 & 0.40 & 0.46 & 0.73 & 0.54 & 0.57 \\
Gemma 3-12B   & 0.44 & 0.71 & 0.58 & 0.57 & 0.54 & 0.80 & 0.55 & 0.63 & 0.33 & 0.58 & 0.32 & 0.41 & 0.46 & 0.73 & 0.51 & 0.57 \\
Gemma 3-27B   & 0.41 & 0.68 & 0.43 & 0.51 & 0.63 & 0.84 & 0.61 & 0.69 & 0.32 & 0.58 & 0.34 & 0.41 & 0.69 & 0.84 & 0.69 & 0.74 \\
\hdashline
LLaMA 3.2-3B  & 0.30 & 0.50 & 0.34 & 0.38 & 0.36 & 0.62 & 0.38 & 0.45 & 0.36 & 0.58 & 0.40 & 0.44 & 0.47 & 0.72 & 0.48 & 0.56 \\
LLaMA 3.1-8B  & 0.32 & 0.53 & 0.34 & 0.40 & 0.40 & 0.64 & 0.43 & 0.49 & 0.40 & 0.61 & 0.44 & 0.48 & 0.55 & 0.76 & 0.58 & 0.63 \\
LLaMA 3.3-70B & 0.42 & 0.67 & 0.46 & 0.52 & 0.43 & 0.72 & 0.52 & 0.56 & 0.45 & 0.67 & 0.47 & 0.53 & 0.58 & 0.81 & 0.64 & 0.67 \\
\hdashline
Qwen 2.5-3B   & 0.34 & 0.58 & 0.42 & 0.45 & 0.39 & 0.70 & 0.47 & 0.52 & 0.34 & 0.59 & 0.38 & 0.44 & 0.45 & 0.75 & 0.56 & 0.59 \\
Qwen 2.5-7B   & 0.33 & 0.63 & 0.46 & 0.48 & 0.39 & 0.73 & 0.48 & 0.53 & 0.37 & 0.62 & 0.41 & 0.47 & 0.46 & 0.76 & 0.49 & 0.57 \\
Qwen 2.5-32B  & 0.40 & 0.66 & 0.50 & 0.52 & 0.37 & 0.73 & 0.48 & 0.52 & 0.39 & 0.65 & 0.47 & 0.50 & 0.41 & 0.70 & 0.46 & 0.53 \\
Qwen 2.5-72B  & 0.33 & 0.61 & 0.41 & 0.45 & 0.47 & 0.77 & 0.50 & 0.58 & 0.38 & 0.65 & 0.42 & 0.49 & 0.43 & 0.73 & 0.42 & 0.52 \\
\bottomrule
\end{tabular}%
}
\caption{Greedy Search CRS Values (temperature = 0.0).}
\label{tab:stylemetrics_crs_greedy_search}
\end{subtable}

\caption{Cross-Response Similarity (CRS) values for AIME, MATH-500, GPQA-Diamond, and GSM8K benchmarks across models, comparing Beam Search (temperature = 1.0) and Greedy Search (temperature = 0.0) generation methods.}
\label{tab:stylemetrics_crs_beam_vs_greedy}
\end{table*}

To assess how instruction styles affect model response characteristics beyond accuracy, we introduce the Cross-Response Similarity (CRS) way of applying RCScore dimensions. This approach (Figure \ref{fig:style_sensitivity_metrics}), detailed in Appendix \ref{sec:algorithm}, quantifies the stylistic consistency among a model's own responses when presented with the same underlying problem framed by different instruction styles. CRS thus highlights a model's tendency to produce stylistically similar or divergent responses based on variations in prompt formulation.
A key aspect of our methodology is the computation of CRS values from the three-dimensional RCScore vector—$\langle \text{Structurality, Lexicality, Coherence} \rangle$—as defined in Appendix \ref{sec:stylemetrics_computation}. This approach retains the multi-dimensional nature of response style and enables fine-grained analysis of which linguistic aspects (syntactic structure, lexical choice, or logical organization) are most affected by instruction variations. For each problem instance, CRS values are computed by aggregating all pairwise comparisons among responses generated from different instruction styles. The detailed aggregation procedure is outlined in Algorithm~\ref{alg:crs_computation_crs_only}.

\subsection{Analyzing Cross-Response Similarity (CRS) with RCScore Dimensions}

We analyze model consistency across instruction styles using the CRS way defined in Section \ref{crs_section}. CRS applies RCScore's three dimensions (Structurality, Lexicality, and Coherence) to quantify response stability. Tables (\ref{tab:stylemetrics_crs_beam_search}, \ref{tab:stylemetrics_crs_greedy_search}) present these values for Beam and Greedy Search respectively, with scores ranging from 0 to 1 (higher values indicating greater consistency).

Our analysis reveals three key patterns. First, decoding strategy significantly affects consistency - Greedy Search consistently yields higher CRS values than Beam Search (e.g., LLaMA 3.3-70B's GSM8K RCScore increases from 0.44 to 0.67). Second, larger models generally demonstrate higher consistency, as seen with LLaMA 3.3-70B outperforming LLaMA 3.2-3B on AIME (0.44 vs. 0.31 with Beam Search). Third, task complexity influences consistency - Gemma 3-27B with Greedy Search achieves 0.51 on AIME but 0.74 on GSM8K. Figure \ref{fig:stylemetrics_all_benchmarks} visually maps these multi-dimensional relationships across models and benchmarks.

\begin{figure*}[t]
  \centering
  \begin{tikzpicture}
  \begin{axis}[
      ybar,
      bar width=10pt,
      enlarge x limits=0.15,
      grid=major,
      grid style={dashed,gray!30},
      ymin=0.30, ymax=0.65,
      ylabel={Overall CRS},
      symbolic x coords={
        {Gemma 4B}, {Gemma 12B}, {Gemma 27B},
        {LLaMA 3B}, {LLaMA 8B}, {LLaMA 70B},
        {Qwen 3B},  {Qwen 7B},  {Qwen 32B}, {Qwen 72B}
      },
      xtick={
        {Gemma 4B}, {Gemma 12B}, {Gemma 27B},
        {LLaMA 3B}, {LLaMA 8B}, {LLaMA 70B},
        {Qwen 3B},  {Qwen 7B},  {Qwen 32B}, {Qwen 72B}
      },
      xticklabel style={
        rotate=45, anchor=east,
        font=\small,
        align=right
      },
      nodes near coords,
      every node near coord/.append style={
        font=\footnotesize,
        /pgf/number format/fixed,
        /pgf/number format/precision=2,
        yshift=-4pt
      },
      legend style={
        at={(0.5,1.02)}, anchor=south,
        legend columns=2,
        font=\small
      },
      width=14cm, height=7cm
    ]

    \addplot+[draw=none,  bar shift=-3pt, forget plot]
      coordinates {({Gemma 4B}, 0.43) ({Gemma 12B}, 0.42) ({Gemma 27B}, 0.45)};
    \addplot+[draw=none,  bar shift=-3pt, forget plot]
      coordinates {({LLaMA 3B}, 0.37) ({LLaMA 8B}, 0.38) ({LLaMA 70B}, 0.46)};
    \addplot+[draw=none,   bar shift=-3pt, forget plot]
      coordinates {({Qwen 3B}, 0.38) ({Qwen 7B}, 0.34) ({Qwen 32B}, 0.37) ({Qwen 72B}, 0.44)};
    \addplot+[draw=none,  fill opacity=0.4, bar shift=+3pt, forget plot]
      coordinates {({Gemma 4B}, 0.53) ({Gemma 12B}, 0.55) ({Gemma 27B}, 0.59)};
    \addplot+[draw=none,  fill opacity=0.4, bar shift=+3pt, forget plot]
      coordinates {({LLaMA 3B}, 0.46) ({LLaMA 8B}, 0.50) ({LLaMA 70B}, 0.57)};
    \addplot+[draw=none,   fill opacity=0.4, bar shift=+3pt, forget plot]
      coordinates {({Qwen 3B}, 0.50) ({Qwen 7B}, 0.51) ({Qwen 32B}, 0.52) ({Qwen 72B}, 0.51)};

    \addlegendimage{ybar, draw=none, fill=gray}
    \addlegendentry{Beam Search}
    \addlegendimage{ybar, draw=none, fill=gray, fill opacity=0.4}
    \addlegendentry{Greedy Search}

  \end{axis}
\end{tikzpicture}
  \vspace{-4pt}
  \caption{Overall CRS values for Gemma, LLaMA, and Qwen model variants. Solid bars correspond to beam search ($T\!=\!1.0$), and semi-transparent bars to greedy search ($T\!=\!0.0$).}
  \label{fig:overall-crs}
\end{figure*}
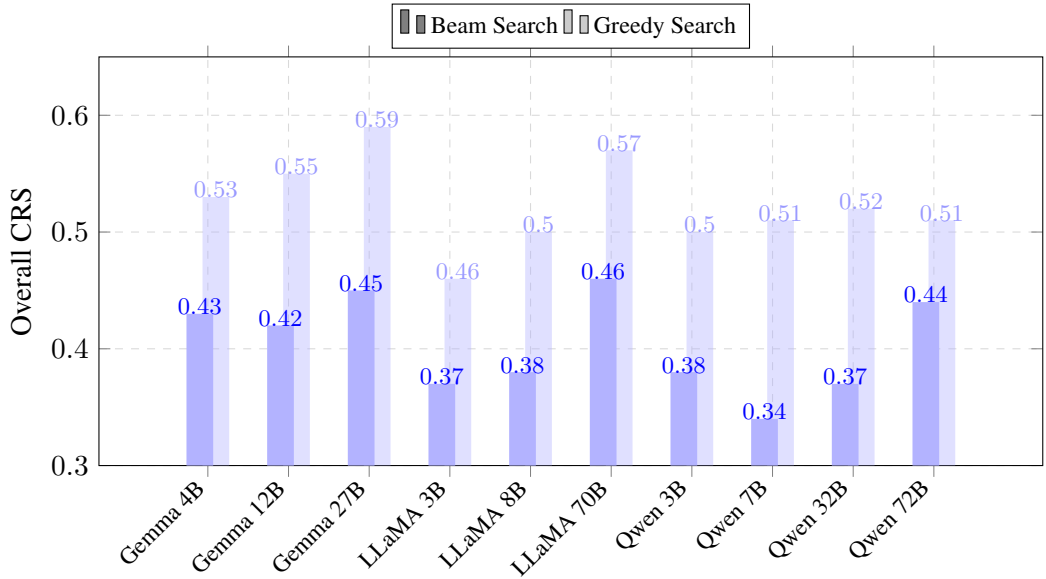

\section{Is Cross-Style Consistency a Reliable Proxy for Accuracy?}

\begin{table}[!h] 
\centering
\resizebox{\columnwidth}{!}{
\begin{tabular}{lrrrr}
\toprule
\textbf{Metric} & \textbf{Pearson} & \textbf{$p$-val} & \textbf{Spearman} & \textbf{$p$-val} \\
 & \textbf{$r$} &  & \textbf{$\rho$} &  \\
\midrule
\multicolumn{5}{l}{\textit{Beam Search (temperature = 1.0)}} \\
RCScore$_{\text{Struct}}$   & 0.57 & $1.4\text{e-}04$ & 0.49 & $0.001$ \\
RCScore$_{\text{Lex}}$      & 0.65 & $5.0\text{e-}06$ & 0.68 & $1.7\text{e-}06$ \\
RCScore$_{\text{Coh}}$       & 0.64 & $9.0\text{e-}06$ & 0.65 & $6.9\text{e-}06$ \\
RCScore$_{\text{Overall}}$         & 0.66 & $3.1\text{e-}06$ & 0.65 & $6.9\text{e-}06$ \\
\midrule
\multicolumn{5}{l}{\textit{Greedy Search (temperature = 0.0)}} \\
RCScore$_{\text{Struct}}$   & 0.675 & $1.74\text{e-}06$ & 0.684 & $1.13\text{e-}06$ \\
RCScore$_{\text{Lex}}$      & 0.790 & $1.39\text{e-}09$ & 0.786 & $1.88\text{e-}09$ \\
RCScore$_{\text{Coh}}$       & 0.656 & $4.38\text{e-}06$ & 0.641 & $8.32\text{e-}06$ \\
RCScore$_{\text{Overall}}$         & 0.733 & $7.43\text{e-}08$ & 0.725 & $1.20\text{e-}07$ \\
\bottomrule
\end{tabular}%
}
\caption{Pearson's $r$ and Spearman's $\rho$ correlations between mean task accuracy and RCScore dimensions (Structurality, Lexicality, Coherence, and Overall) applied through the Cross-Response Similarity (CRS) way across all model-benchmark pairs (N=40) for Beam Search and Greedy Search}

\label{tab:correlation_metrics_crs_beam_greedy}
\end{table}

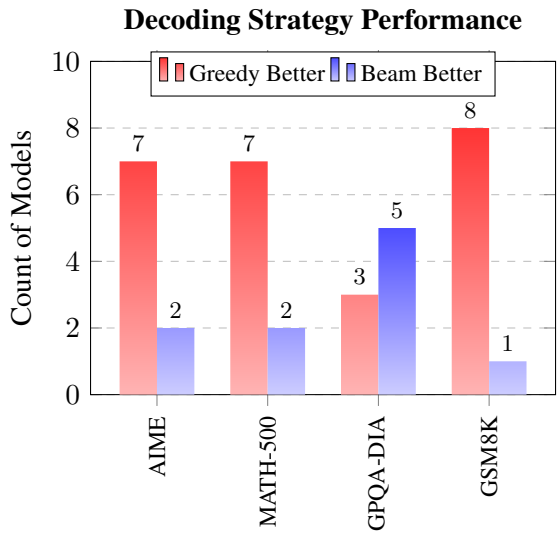
\begin{figure}[!h]
\centering
\begin{tikzpicture}
\begin{axis}[
    title={\textbf{Decoding Strategy Performance}},
    xlabel={},
    symbolic x coords={AIME, MATH-500, GPQA-DIA, GSM8K},
    xtick=data,
    xticklabel style={rotate=90, anchor=east, font=\small},
    width=\columnwidth,
    height=6cm,
    ybar=0pt, 
    bar width=14pt,
    ymin=0,
    ymax=10, 
    enlarge x limits=0.2,
    legend style={at={(0.5,1.03)}, anchor=north, legend columns=2, font=\small},
    ylabel={Count of Models},
    nodes near coords,
    every node near coord/.append style={font=\small},
    ymajorgrids=true,
    grid style=dashed,
]
\addplot[draw=none, top color=red!80, bottom color=red!30] coordinates { 
    (AIME,7)
    (MATH-500,7)
    (GPQA-DIA,3)
    (GSM8K,8)
};
\addplot[draw=none, top color=blue!70, bottom color=blue!20] coordinates { 
    (AIME,2)
    (MATH-500,2)
    (GPQA-DIA,5)
    (GSM8K,1)
};
\legend{Greedy Better, Beam Better}
\end{axis}
\end{tikzpicture}
\caption{Decoding strategy performance: Number of models where Greedy Search or Beam Search achieved higher average accuracy per benchmark. Greedy Search generally performed better, aligning with its higher CRS values (Figure~\ref{fig:overall-crs}). Detailed average accuracy scores are available in Table~\ref{tab:beam_vs_greedy_comparison}.}
\label{fig:greedy_vs_beam_stability_updated}
\end{figure}

To investigate whether a model's consistency across varied instruction styles correlates with its task-solving accuracy, we analyzed the relationship between our CRS values and conventional accuracy metrics. For each of the ten models across four benchmarks (AIME, GSM8K, MATH-500, GPQA), we computed CRS values using RCScore's three dimensions (Structurality, Lexicality, Coherence) and the aggregated RCScore. Concurrently, we calculated a mean accuracy score for each of the 40 model-benchmark pairs by averaging accuracies achieved under each of the four instruction styles. We then measured Pearson's $r$ and Spearman's $\rho$ correlations between these mean accuracies and each dimension of CRS to assess both linear and monotonic relationships.

The results, presented in Table~\ref{tab:correlation_metrics_crs_beam_greedy}, reveal a significant positive correlation between a model's ability to maintain stylistic consistency and its overall task performance. Notably, the aggregated RCScore and its Lexicality dimension when applied through the CRS way showed strong correlations with mean accuracy. For instance, Lexicality exhibited a Pearson's $r > 0.65$ under beam search and $r \approx 0.79$ under greedy search. This suggests that models producing lexically consistent explanations across different instruction styles tend to achieve higher accuracy. More broadly, all dimensions demonstrated statistically significant positive correlations with task accuracy when measured via the CRS approach. This consistent pattern indicates that higher cross-style self-similarity in a model's responses is a robust indicator of better task performance. These findings underscore the utility of RCScore dimensions applied through CRS in capturing nuanced behavioral traits of LLMs—specifically, the ability to generate consistent outputs irrespective of instruction style—that are meaningfully aligned with their fundamental task-solving capabilities. Thus, cross-style consistency, as quantified through RCScore dimensions, emerges not just as a measure of stylistic stability but as a valuable proxy for a model's underlying accuracy and reliability.

\subsection{Parameter-Scale and Decoding Strategy Effects on CRS}
Figure~\ref{fig:overall-crs} displays CRS values, highlighting the influence of model scale and decoding strategy on stylistic consistency. Larger models (>70B), such as LLaMA 3.3-70B and Qwen 2.5-72B, generally demonstrate higher CRS, suggesting that increased parameter scale improves stylistic stability against varied instruction phrasings. Decoding strategy also plays a critical role: greedy search ($T=0.0$) consistently yields 7-13\% higher CRS scores than beam search ($T=1.0$) across all models, indicating deterministic decoding produces more stylistically stable outputs. Notably, this enhanced consistency with greedy search correlates with improved task performance. As illustrated in Figure~\ref{fig:greedy_vs_beam_stability_updated}, greedy decoding led to higher average accuracy for a majority of models on the AIME, MATH-500, and GSM8K benchmarks. While Beam Search appears to have a slight edge on GPQA-Diamond for more models, the overall low accuracy scores on this benchmark (typically below 10\% for most models, as seen in Table~\ref{tab:combined_accuracy_by_style}) suggest that these particular differences in decoding strategy performance might be less indicative of a general trend. Overall, the deterministic approach of greedy search also narrows the CRS performance gap between smaller and larger models, particularly benefiting smaller models in maintaining consistency.

\section{Conclusion}
We introduce RCScore, a multi-dimensional metrics designed to assess LLM sensitivity to instruction style, extending traditional accuracy-focused evaluations. RCScore quantifies response variations along three core dimensions: Structurality, Lexicality, and Coherence. Through experiments applying varied instruction styles to established benchmarks, we demonstrate that LLM performance can fluctuate based on instruction phrasing. Our findings introduce Cross-Response Similarity (CRS) as a measure of stylistic self-consistency, establishing a strong positive correlation between CRS and task accuracy. Additionally, we observe that larger models exhibit greater stylistic stability, and deterministic decoding produces more consistent outputs. These insights suggest that cross-style consistency serves as a valuable indicator of model reliability, offering deeper perspectives on LLM robustness.

\section{Limitations}
While RCScore offers a more nuanced evaluation of instruction style sensitivity, its current implementation relies on a predefined set of four linguistic styles and specific English-language NLP tools for dimensional analysis (e.g., dependency parsing, TF-IDF). The framework's applicability to other languages or a broader range of stylistic variations (e.g., code-switching, highly informal language) remains to be explored, and the computational cost, though minimized, might still be a factor for extremely large-scale evaluations across numerous model-benchmark pairs. 

Additionally, our focus on mathematical and reasoning tasks, while methodologically sound, may not generalize to more subjective domains like creative writing or emotional support, where style sensitivity might manifest differently. The current RCScore dimensions also lack validation through human evaluation to confirm their alignment with human perceptions of consistency or quality. Furthermore, the weights for aggregating RCScore dimensions are currently uniform, and future work could investigate task-dependent or empirically derived weighting schemes.

\section{Ethical Consideration}
The primary ethical considerations for RCScore relate to its potential application and interpretation. While designed to improve LLM evaluation by revealing sensitivities, there is a risk that findings could be used to "over-optimize" models for specific instruction styles, potentially at the expense of generalizability or by inadvertently encoding biases present in the chosen stylistic variations. 

The datasets used (AIME, MATH, GPQA, GSM8K) are standard academic benchmarks, and the instruction variations are syntactically derived without introducing new factual content, minimizing the risk of generating harmful or biased test material. We acknowledge that inconsistent performance across instruction styles raises accessibility concerns, as users with different cultural backgrounds or neurodivergent conditions might naturally prefer specific formulation patterns, potentially creating uneven experiences with AI systems. 

Additionally, the increased computational requirements for multi-style evaluation introduce environmental considerations that should be balanced against the benefits of more comprehensive assessment. We encourage users of RCScore to be mindful of these aspects and to employ the framework as a tool for understanding and improving model robustness broadly, rather than for narrow stylistic optimization.

\bibliography{custom}
\appendix

\section{RCScore Computation Details}
\label{sec:stylemetrics_computation}

\subsection{Dimension Formulations}
All three metrics employ a unified computational approach where similarity scores are calculated between document pairs, normalized to the [0,1] range, with higher values indicating greater similarity along the respective dimension.

\subsubsection{Structurality}
We compute Structurality through syntactic pattern comparison between aligned sentence pairs:

\begin{align}
Struct(D_a, D_b) &= \frac{1}{|M|}\sum_{(s_i, t_i) \in M} J(P(s_i), P(t_i))
\end{align}

where $D_a, D_b$ denote the source and target documents, $M = \{(s_1,t_1), (s_2,t_2), \ldots, (s_n,t_n)\}$ represents the set of semantically aligned sentence pairs determined via BERTScore similarity, $P(s)$ extracts the set of syntactic patterns from sentence $s$, and $J(A,B) = \frac{|A \cap B|}{|A \cup B|}$ calculates the Jaccard similarity between pattern sets. Each syntactic pattern has the form $\langle pos_t, dep_r, pos_h \rangle$ representing the part-of-speech tag, dependency relation, and head token's part-of-speech.

This approach captures fine-grained syntactic relationships while abstracting away from specific lexical choices, allowing us to measure structural similarity even when vocabulary differs substantially.

\subsubsection{Lexicality}
We define Lexicality as the weighted sum of two complementary similarity measures:

\begin{align}
\text{Lex}(D_a, D_b) = w_{\text{TF}} \cdot S_{\text{TF}} + w_{\text{RL}} \cdot S_{\text{RL}}
\end{align}

where $S_{\text{TF}}$ represents the TF-IDF cosine similarity between document vectors, and $S_{\text{RL}}$ quantifies the ROUGE-L F-measure based on longest common subsequence. $w_{\text{TF}}$ and $w_{\text{RL}}$ are weights with default values $w_{\text{TF}} = w_{\text{RL}} = 0.5.$ This combined approach balances global vocabulary distribution similarity (TF-IDF) with local sequential word matching (ROUGE-L).

\subsubsection{Coherence}
We formulate Coherence as the product of structural alignment and content similarity:

\begin{equation}
\text{Coherence}(D_a, D_b) = S(D_a, D_b) \cdot C_w(D_a, D_b)
\end{equation}

where $S(D_a, D_b) = w_O \cdot O + w_P \cdot P + w_N \cdot N + w_C \cdot C_s$ represents the structural similarity score combining four components: (1) $O = \frac{\tau + 1}{2}$, the normalized Kendall's Tau correlation ($\tau$) of chunk order; (2) $P$, position matching with emphasis on document endpoints; (3) $N$, sequential continuity between matched chunks; and (4) $C_s$, semantic similarity between aligned chunks. The weights $w_O, w_P, w_N, w_C$ have default values of 0.25 each. The term $C_w(D_a, D_b) = C_s^2$ applies a quadratic content-weighted penalty to ensure that structural similarities are only meaningful when the compared chunks contain related information.

The computation involves segmenting documents into semantic chunks of adaptive size $k$, aligning chunks using a combined BERTScore and TF-IDF similarity matrix, and analyzing the sequence and position of matched chunk pairs.

\begin{algorithm}
\caption{Applying RCScore Dimensions through the CRS Way}
\label{alg:crs_computation_crs_only}
\begin{algorithmic}[1]
\Require Style-variant responses $\{R_s\}$ where $s \in S$
\Ensure Three-dimensional CRS values using RCScore dimensions

\Function{RCScore}{$D_1, D_2$}
    \State \Return $\langle \text{Structurality}, \text{Lexicality},$ \\
    \hskip\algorithmicindent\hskip\algorithmicindent $\text{Coherence} \rangle$ \Comment{3D similarity vector}
\EndFunction

\State $\text{CRS}_{\text{vectors}} \gets \emptyset$
\State $n \gets \binom{|S|}{2}$ \Comment{Number of style pairs}
\For{$(s_i, s_j) \in \binom{S}{2}$} \Comment{All style pairs}
    \State $\vec{sim}_{i,j} \gets \text{RCScore}(R_{s_i}, R_{s_j})$
    \State $\text{CRS}_{\text{vectors}} \gets \text{CRS}_{\text{vectors}} \cup \{\vec{sim}_{i,j}\}$
\EndFor

\State $\text{CRS} \gets \frac{1}{n}\sum_{\vec{v} \in \text{CRS}_{\text{vectors}}} \vec{v}$ \Comment{Dimension-preserving average}

\State \Return $\text{CRS} = \langle \text{CRS}_{\text{Struct}}, \text{CRS}_{\text{Lex}},$ \\
\hskip\algorithmicindent\hskip\algorithmicindent $\text{CRS}_{\text{Coh}} \rangle$
\end{algorithmic}
\end{algorithm}

\subsection{Cross-Response Similarity (CRS) Way of Applying RCScore}

\label{sec:algorithm}

\begin{table*}[t!]
\centering

\begin{subtable}{\textwidth}
\centering
\resizebox{\textwidth}{!}{%
\begin{tabular}{|l|cccc|cccc|cccc|cccc|}
\hline
\textbf{Model} & \multicolumn{4}{c|}{\textbf{AIME}} & \multicolumn{4}{c|}{\textbf{MATH-500}} & \multicolumn{4}{c|}{\textbf{GPQA-Diamond}} & \multicolumn{4}{c|}{\textbf{GSM8K}} \\
& \textbf{S1} & \textbf{S2} & \textbf{S3} & \textbf{S4} & \textbf{S1} & \textbf{S2} & \textbf{S3} & \textbf{S4} & \textbf{S1} & \textbf{S2} & \textbf{S3} & \textbf{S4} & \textbf{S1} & \textbf{S2} & \textbf{S3} & \textbf{S4} \\
\hline
Gemma 3-4B & 6.7 & 10.0 & 6.7 & 3.3 & 65.8 & 64.0 & 66.0 & 67.0 & 4.5 & 3.0 & 4.0 & 2.5 & 88.3 & 87.9 & 88.9 & 89.2 \\
Gemma 3-12B & 23.3 & 23.3 & 23.3 & 20.0 & 71.4 & 71.6 & 73.0 & 72.4 & 4.5 & 4.0 & 6.6 & 6.6 & 92.0 & 91.7 & 92.3 & 91.9 \\
Gemma 3-27B & 30.0 & 26.7 & 33.3 & 20.0 & 75.8 & 77.6 & 76.8 & 77.4 & 6.1 & 7.1 & 6.1 & 6.6 & 94.8 & 93.9 & 94.6 & 94.1 \\
\hdashline
LLaMA 3.2-3B & 3.3 & 0.0 & 6.7 & 10.0 & 41.8 & 41.6 & 41.0 & 39.8 & 5.1 & 2.0 & 4.0 & 5.1 & 78.3 & 76.4 & 71.8 & 78.4 \\
LLaMA 3.1-8B & 3.3 & 10.0 & 6.7 & 6.7 & 42.8 & 43.6 & 42.4 & 46.4 & 2.0 & 3.5 & 4.5 & 2.5 & 82.1 & 82.9 & 84.2 & 83.5 \\
LLaMA 3.3-70B & 20.0 & 23.3 & 26.7 & 23.3 & 64.4 & 64.4 & 61.6 & 63.6 & 6.1 & 8.6 & 5.6 & 5.6 & 90.4 & 90.1 & 91.4 & 90.8 \\
\hdashline
Qwen 2.5-3B & 3.3 & 3.3 & 3.3 & 6.7 & 55.0 & 54.2 & 55.2 & 55.4 & 2.5 & 2.5 & 3.0 & 3.5 & 83.3 & 79.1 & 84.3 & 83.3 \\
Qwen 2.5-7B & 13.3 & 6.7 & 13.3 & 13.3 & 64.0 & 63.0 & 58.2 & 65.0 & 5.6 & 3.5 & 4.5 & 4.0 & 87.1 & 88.3 & 82.5 & 86.2 \\
Qwen 2.5-32B & 13.3 & 13.3 & 16.7 & 16.7 & 69.8 & 68.8 & 67.8 & 69.8 & 3.5 & 2.0 & 6.1 & 2.0 & 92.6 & 91.4 & 93.2 & 93.0 \\
Qwen 2.5-72B & 30.0 & 23.3 & 13.3 & 23.3 & 70.8 & 69.6 & 71.2 & 71.2 & 4.0 & 6.6 & 5.6 & 5.1 & 92.3 & 92.9 & 92.9 & 92.5 \\
\hline
\end{tabular}%
}
\caption{Beam search (temperature = 1.0).}
\label{tab:accuracy_by_style_temp1}
\end{subtable}

\vspace{1em} 

\begin{subtable}{\textwidth}
\centering
\resizebox{\textwidth}{!}{%
\begin{tabular}{|l|cccc|cccc|cccc|cccc|}
\hline
\textbf{Model} & \multicolumn{4}{c|}{\textbf{AIME}} & \multicolumn{4}{c|}{\textbf{MATH-500}} & \multicolumn{4}{c|}{\textbf{GPQA-Diamond}} & \multicolumn{4}{c|}{\textbf{GSM8K}} \\
& \textbf{S1} & \textbf{S2} & \textbf{S3} & \textbf{S4} & \textbf{S1} & \textbf{S2} & \textbf{S3} & \textbf{S4} & \textbf{S1} & \textbf{S2} & \textbf{S3} & \textbf{S4} & \textbf{S1} & \textbf{S2} & \textbf{S3} & \textbf{S4} \\
\hline
Gemma 3-4B & 13.3 & 13.3 & 6.7 & 10.0 & 66.6 & 65.8 & 64.6 & 64.6 & 3.5 & 2.5 & 3.0 & 2.0 & 87.1 & 84.8 & 84.8 & 86.1 \\
Gemma 3-12B & 3.3 & 3.3 & 3.3 & 3.3 & 52.4 & 52.6 & 54.6 & 50.8 & 6.1 & 6.6 & 3.0 & 5.1 & 90.0 & 89.9 & 90.3 & 91.0 \\
Gemma 3-27B & 26.7 & 30.0 & 30.0 & 23.3 & 76.2 & 76.4 & 76.8 & 78.0 & 6.1 & 5.6 & 5.1 & 6.6 & 94.5 & 94.2 & 94.2 & 94.6 \\
\hdashline
LLaMA 3.2-3B & 10.0 & 13.3 & 13.3 & 10.0 & 41.6 & 42.6 & 42.4 & 45.4 & 3.0 & 4.0 & 2.5 & 5.1 & 79.8 & 79.1 & 78.0 & 79.8 \\
LLaMA 3.1-8B & 6.7 & 3.3 & 10.0 & 10.0 & 43.2 & 45.0 & 43.6 & 44.8 & 3.5 & 5.1 & 3.5 & 3.0 & 84.7 & 85.2 & 84.9 & 84.9 \\
LLaMA 3.3-70B & 26.7 & 23.3 & 30.0 & 26.7 & 64.4 & 63.0 & 61.8 & 65.0 & 7.1 & 8.1 & 7.1 & 7.1 & 93.7 & 93.3 & 93.2 & 93.3 \\
\hdashline
Qwen 2.5-3B & 6.7 & 13.3 & 10.0 & 3.3 & 57.4 & 55.2 & 58.2 & 57.2 & 1.5 & 1.5 & 3.0 & 3.0 & 84.7 & 84.8 & 84.2 & 84.6 \\
Qwen 2.5-7B & 20.0 & 16.7 & 10.0 & 16.7 & 66.2 & 66.6 & 65.8 & 64.6 & 2.0 & 2.0 & 3.0 & 1.5 & 90.5 & 88.9 & 89.4 & 89.3 \\
Qwen 2.5-32B & 16.7 & 10.0 & 20.0 & 20.0 & 71.6 & 70.6 & 72.2 & 69.0 & 5.1 & 6.6 & 7.1 & 4.0 & 92.8 & 92.8 & 92.3 & 92.7 \\
Qwen 2.5-72B & 20.0 & 26.7 & 20.0 & 20.0 & 72.2 & 70.8 & 71.8 & 69.6 & 5.6 & 3.5 & 6.1 & 5.1 & 92.8 & 92.3 & 93.0 & 93.3 \\
\hline
\end{tabular}%
}
\caption{Greedy search (temperature = 0.0).}
\label{tab:accuracy_by_style_temp0}
\end{subtable}

\caption{Accuracy by Instruction Style (S1: Declarative, S2: Interrogative, S3: Exclamative, S4: Imperative) on four benchmarks (AIME, MATH-500, GPQA-Diamond, and GSM8K) for different decoding strategies.}
\label{tab:combined_accuracy_by_style}
\end{table*}

To quantify how instruction styles influence model responses, we apply RCScore dimensions through the Cross-Response Similarity (CRS) way (Algorithm \ref{alg:crs_computation_crs_only}):

\begin{itemize}
    \item \textbf{Cross-Response Similarity (CRS)} - measures the consistency across $\binom{|S|}{2}$ pairs of responses generated under different instruction formulations using RCScore dimensions.
\end{itemize}

The key innovation in our approach is preserving RCScore's three-dimensional nature when calculating cross-style consistency. Each comparison produces a vector $\langle struct, lex, coh \rangle$ representing Structurality, Lexicality, and Coherence dimensions. This allows for a detailed analysis of which specific aspects of reasoning style are most affected by instruction variations.

For each problem, we compute:
\begin{align}
\text{CRS}_p &= \text{Aggregate}\{\vec{sim}_{i,j} | (s_i, s_j) \in \binom{S}{2}\}
\end{align}

where $S$ is the set of instruction styles, and $\vec{sim}_{i,j}$ is the similarity vector between responses generated under instruction styles $s_i$ and $s_j$, calculated using RCScore dimensions.

\begin{table*}[t]
\centering
\small
\begin{tabular}{|l|cccc|cccc|}
\hline
\multirow{2}{*}{\textbf{Model}} & \multicolumn{4}{c|}{\textbf{Style Sensitivity Index (SSI) at temp=1.0}} & \multicolumn{4}{c|}{\textbf{Style Sensitivity Index (SSI) at temp=0.0}} \\
& \textbf{AIME} & \textbf{MATH} & \textbf{GPQA} & \textbf{GSM8K} & \textbf{AIME} & \textbf{MATH} & \textbf{GPQA} & \textbf{GSM8K} \\
\hline
Gemma 3-4B    & 2.11 & 0.18 & 0.95 & 0.07 & 1.30 & 0.14 & 1.07 & 0.12 \\
Gemma 3-12B   & 0.52 & 0.11 & 1.08 & 0.03 & 0.00 & 0.28 & 1.20 & 0.05 \\
Gemma 3-27B   & 1.55 & 0.09 & 0.43 & 0.04 & 0.99 & 0.09 & 0.59 & 0.02 \\
\hline
LLaMA 3-3B    & 4.23 & 0.19 & 1.20 & 0.40 & 0.92 & 0.34 & 1.05 & 0.10 \\
LLaMA 3-8B    & 2.39 & 0.33 & 1.12 & 0.10 & 2.49 & 0.16 & 0.81 & 0.03 \\
LLaMA 3-70B   & 0.86 & 0.16 & 0.65 & 0.06 & 0.85 & 0.18 & 0.25 & 0.02 \\
\hline
Qwen 2.5-3B   & 2.26 & 0.09 & 0.79 & 0.25 & 2.84 & 0.18 & 1.78 & 0.03 \\
Qwen 2.5-7B   & 1.49 & 0.43 & 0.61 & 0.27 & 1.54 & 0.12 & 1.24 & 0.08 \\
Qwen 2.5-32B  & 0.70 & 0.12 & 2.67 & 0.08 & 1.29 & 0.15 & 1.01 & 0.03 \\
Qwen 2.5-72B  & 2.42 & 0.09 & 0.67 & 0.03 & 1.07 & 0.14 & 0.85 & 0.04 \\
\hline
\end{tabular}
\caption{\textbf{Style sensitivity across tasks and temperature settings.} The Style Sensitivity Index (SSI) quantifies variation in model performance across instruction styles, calculated as $\text{SSI} = 5 \cdot (\sigma/\mu) + 0.05 \cdot (\max-\min)$, where $\sigma$ is population standard deviation and $\mu$ is mean accuracy. Higher values indicate greater performance variability. AIME exhibits the highest sensitivity (average SSI: 1.83 at temp=1.0), followed by GPQA-Diamond (average SSI: 1.02), while MATH-500 and GSM8K show lower sensitivity (average SSI: 0.18 and 0.13). While temperature=0.0 generally reduces style sensitivity, three models (Qwen 2.5-3B/32B and LLaMA 3-8B) exhibit increased sensitivity at temperature=0.0 on AIME, demonstrating that deterministic generation can sometimes amplify instruction style effects.}
\label{tab:all_benchmarks_sensitivity_final} 
\end{table*}

\section{Experimental Implementation Details}
\label{sec:experimental_implementation}
\begin{figure}[!ht]
\centering
\begin{tcolorbox}[
    width=0.5\textwidth,
    colback=blue!5,
    colframe=black!40,
    boxrule=0.5pt,
    arc=1mm,
    fontupper=\footnotesize,
    left=5pt,
    right=5pt,
    top=5pt,
    bottom=5pt,
    boxsep=3pt,
]
\begin{tabularx}{\textwidth}{>{\raggedright\arraybackslash}X}
\textbf{\textcolor{black!80}{Prompt Template for Instruction Style Variations}} \\
\midrule
\rowcolor{blue!10} \textbf{Code Implementation:} \\
\midrule
\textbf{\textcolor{blue!70!black}{Instruction Type Definition:}} \\
\texttt{sentence\_type = [} \\
\quad \textcolor{blue!70!black}{\# Declarative} \\
\quad \texttt{"The problem should be solved step by step. The answer is to be} \\
\quad \texttt{suggested in the following format."} \\
\quad \textcolor{violet!80!black}{\# Interrogative} \\
\quad \texttt{"Could you solve the problem step by step? Would you suggest} \\
\quad \texttt{the answer in the following format?"} \\
\quad \textcolor{green!60!black}{\# Exclamative} \\
\quad \texttt{"How important it is to solve the problem step by step! What a} \\
\quad \texttt{necessity it is to suggest the answer in the following format!"} \\
\quad \textcolor{orange!80!black}{\# Imperative} \\
\quad \texttt{"Solve the problem step by step. Suggest the answer in the} \\
\quad \texttt{following format."} \\
\texttt{]} \\
\midrule
\textbf{\textcolor{blue!70!black}{Message Structure:}} \\
\texttt{messages = [\{} \\
\quad \texttt{"role": "user",} \\
\quad \texttt{"content": (} \\
\quad\quad \texttt{f"\{text\}\textbackslash n\textbackslash n"} \\
\quad\quad \texttt{f"\{sentence\_type\}\textbackslash n"} \\
\quad\quad \texttt{f"Solution: [explanation]\textbackslash n"} \\
\quad\quad \texttt{f"Answer: [answer]"} \\
\quad \texttt{)} \\
\texttt{\}]} \\
\end{tabularx}
\end{tcolorbox}
\caption{\textbf{Prompt implementation for RCScore experiments.} We programmatically append different instruction styles to each benchmark problem while maintaining a consistent output structure. This approach enables efficient evaluation of style sensitivity without modifying the core problem content.}
\label{fig_prompt_template}
\end{figure}

RCScore evaluates model performance across different instruction styles while maintaining consistent semantic content. Figure \ref{fig_prompt_template} shows our implementation approach for applying these style variations.

Rather than completely rephrasing benchmark problems, which would be computationally expensive and potentially introduce unintended semantic shifts, we adopted a lightweight approach that appends style-specific instructions to the original problem text. This preserves the core problem content while varying only the instruction component.

As shown in Figure \ref{fig_prompt_template}, our implementation defines four instruction styles—Declarative, Interrogative, Exclamative, and Imperative—following linguistic clause type classifications. Each style maintains consistent main verbs (solve, suggest) while varying syntactic structure. We carefully controlled the Type-Token Ratio (0.75-0.78) across all styles to ensure lexical complexity remained comparable.

This implementation approach enables efficient evaluation of style sensitivity at scale across multiple benchmarks and models, with minimal disruption to the original benchmark problems.

\section{Detailed Performance Results}
\subsection{Performance with Beam Search (Temperature = 1.0)}
Table \ref{tab:accuracy_by_style_temp1} provides comprehensive accuracy results for all evaluated models across the four instruction styles (S1: Declarative, S2: Interrogative, S3: Exclamative, S4: Imperative) and four reasoning tasks when using beam search with \textbf{temperature = 1.0}. These detailed measurements complement the visualization in Figure \ref{fig:combined_style_sensitivity} (a), providing exact numerical values.

The results reveal several notable patterns. Within each model family, larger parameter counts generally correlate with higher accuracy. For example, across most tasks, Gemma 3-27B shows higher accuracy than Gemma 3-4B. Different benchmarks exhibit varying levels of style sensitivity; AIME, for instance, shows considerable variance (e.g., Qwen 2.5-72B ranges from 13.3\% with S3 to 30.0\% with S1, a 16.7 percentage point (pp) difference). In contrast, on GSM8K, the variation for many models is smaller, though LLaMA 3.2-3B still shows a 6.6\% range (71.8\% with S3 to 78.4\% with S4).

Model families demonstrate distinct patterns of style preference, which can also vary by task. For example, on AIME, Gemma 3-27B achieved its highest accuracy (33.3\%) with the Exclamative style (S3), while Qwen 2.5-72B performed best with the Declarative style (S1) at 30.0\%. The standard deviation of performance across styles can serve as an indicator of style robustness, with lower deviations suggesting more consistent performance.

\subsection{Performance with Greedy Search (Temperature = 0.0)}
Table \ref{tab:accuracy_by_style_temp0} presents accuracy results for the same set of models and tasks when using greedy search (\textbf{temperature = 0.0}), providing a deterministic generation scenario. These results are visualized in Figure \ref{fig:combined_style_sensitivity} (b). Similar to beam search results, larger models within each family generally show higher accuracy (e.g., Gemma 3-27B and LLaMA 3.3-70B typically outperform their smaller counterparts).

Benchmarks continue to show varying degrees of sensitivity to instruction styles even in this deterministic setting. AIME and GPQA-Diamond tend to exhibit more pronounced style-based variations compared to GSM8K, where performance is often more consistent. For instance, on AIME, LLaMA 3.1-8B's accuracy ranged from 3.3\% (S2) to 10.0\% (S3/S4), a 6.7\% difference. On GPQA-Diamond, Qwen 2.5-32B varied from 4.0\% (S4) to 7.1\% (S3), a 3.1\% difference. Distinct style preferences persist, with models showing varying optimal styles depending on the task and model size. Standard deviations across styles are generally smaller than at temperature = 1.0, indicating more consistent performance in deterministic generation.

\subsection{Comparative Analysis of Temperature Effects}
Comparing model performance with beam search (temperature = 1.0) versus greedy search (temperature = 0.0) reveals important insights. Models generally exhibit reduced cross-style variance with greedy search. For example, on AIME, the maximum observed accuracy difference for LLaMA 3.1-8B was 6.7\% with greedy search, compared to larger gaps such as 16.7\% for Qwen 2.5-72B under beam search.

Despite this reduction in variance, significant style-based performance differences persist even in deterministic generation. As noted, Qwen 2.5-32B on GPQA-Diamond still exhibited a 3.1\% difference with greedy search. While greedy search tends to yield more consistent performance across styles for most models, it does not always result in optimal accuracy. In several instances, models achieve their highest accuracy on particular style-task combinations with beam search at temperature = 1.0 (e.g., Gemma 3-27B on AIME with S3: 33.3\% at temp=1.0 vs. 30.0\% at temp=0.0; LLaMA 3.3-70B on GPQA-Diamond with S2: 8.6\% at temp=1.0 vs. 8.1\% at temp=0.0), suggesting that some sampling diversity can be beneficial.

These findings indicate that practitioners should consider both instruction style and temperature settings. For applications requiring maximum consistency, greedy search with carefully chosen instruction styles may be preferable. Conversely, applications seeking peak performance might benefit from temperature > 0 with task-specific style optimization. The persistence of style-based variation across temperature settings underscores the importance of comprehensive evaluation frameworks like RCScore.

\section{Cross-Task Patterns of Style Sensitivity}
\label{sec:style_sensitivity_analysis}

To quantify how instruction style affects model performance across tasks and temperature settings, we developed the Style Sensitivity Index (SSI):

\begin{equation}
\text{SSI} = 5 \cdot \frac{\sigma}{\mu} + 0.05 \cdot (\max-\min)
\end{equation}

where $\sigma$ is the standard deviation of accuracy across styles, $\mu$ is mean accuracy, and $(\max-\min)$ captures the absolute performance range. This metric integrates both relative variation and absolute performance differences.

Table~\ref{tab:all_benchmarks_sensitivity_final} presents SSI values for all models across four benchmarks at temperatures 1.0 and 0.0. Style sensitivity patterns persist across all tasks but with task-dependent magnitudes: AIME exhibits highest sensitivity (avg SSI: 1.83 at temp=1.0), followed by GPQA-Diamond (avg SSI: 1.02), with MATH-500 and GSM8K showing substantially lower sensitivity (avg SSI: 0.18 and 0.13). This hierarchy suggests that task complexity and open-endedness amplify instruction phrasing effects.

\subsection{Analysis of Style Sensitivity on AIME}
AIME demonstrates exceptionally high style sensitivity, with several models showing SSI values exceeding 2.0 at temperature=1.0. LLaMA 3B exhibits the highest sensitivity (SSI: 4.23), followed by LLaMA 8B (2.39), Qwen 72B (2.42), and Qwen 3B (2.26). The competitive nature of AIME problems, requiring advanced mathematical reasoning and precise numerical solutions, appears particularly vulnerable to instruction formulation variations.

While temperature=0.0 generally reduces sensitivity, significant variations persist in deterministic generation. Notably, three models exhibit increased sensitivity at temperature=0.0: Qwen 3B (2.26→2.84), Qwen 32B (0.70→1.29), and LLaMA 8B (2.39→2.49). This contradicts the expectation that deterministic generation would universally stabilize performance across instruction styles.

\subsection{Analysis of Style Sensitivity on MATH-500}
MATH-500 shows consistently lower style sensitivity than AIME despite both being mathematical reasoning tasks. SSI values predominantly remain below 0.2, with Qwen 7B at temperature=1.0 being the notable exception (SSI: 0.43). The structured format of MATH-500 problems likely contributes to this reduced sensitivity, providing consistent mathematical notation that models can process regardless of instruction style.

Temperature effects on MATH-500 sensitivity vary by model. Qwen 7B shows significantly higher sensitivity at temperature=1.0 (0.43) compared to temperature=0.0 (0.12), while Gemma 12B exhibits the opposite pattern (0.11→0.28). LLaMA 3B demonstrates significant sensitivity at both temperature settings (0.19→0.34).

\subsection{Analysis of Style Sensitivity on GPQA-Diamond}
GPQA-Diamond exhibits moderate to high style sensitivity with diverse patterns across model families. Qwen 32B shows exceptionally high sensitivity at temperature=1.0 (SSI: 2.67), followed by LLaMA 3B (1.20) and LLaMA 8B (1.12). The scientific reasoning required by GPQA-Diamond appears particularly susceptible to instruction variations, especially for certain model architectures.

Temperature effects on GPQA sensitivity show model-specific patterns. While LLaMA 70B demonstrates reduced sensitivity at temperature=0.0 (0.65→0.25), several models exhibit increased sensitivity: Qwen 3B (0.79→1.78), Qwen 7B (0.61→1.24), and Gemma 12B (1.08→1.20).

\subsection{Analysis of Style Sensitivity on GSM8K}
GSM8K demonstrates the lowest style sensitivity among all benchmarks, with most models exhibiting SSI values below 0.1, particularly at temperature=0.0. The well-structured nature of elementary arithmetic word problems likely contributes to this reduced sensitivity, presenting unambiguous reasoning paths regardless of instruction phrasing.

Temperature significantly impacts GSM8K style sensitivity. At temperature=1.0, smaller models show moderate sensitivity: LLaMA 3B (SSI: 0.40), Qwen 7B (0.27), and Qwen 3B (0.25). However, at temperature=0.0, nearly all models demonstrate minimal sensitivity (SSI < 0.05), with only LLaMA 3B (0.10) and Gemma 4B (0.12) showing SSI values above 0.1.

\definecolor{mygreen}{rgb}{0.1, 0.5, 0.1}
\definecolor{myred}{rgb}{0.7, 0.1, 0.1}
\definecolor{myblue}{rgb}{0.1, 0.1, 0.7}
\begin{table*}[!htbp]
\centering
\resizebox{\textwidth}{!}{%
\begin{tabular}{p{0.23\textwidth} p{0.47\textwidth} r r r r l}
\toprule
\textbf{Reference} & \textbf{Comparison} & \textbf{Struct} & \textbf{Lexi} & \textbf{Cohe} & \textbf{RCScore} & \textbf{Similarity} \\
\midrule
\multirow{10}{=}{
The analysis of algorithmic complexity \textcolor{myblue}{requires a systematic approach}. \textcolor{myblue}{First, we must identify the basic operations} within the algorithm. \textcolor{myblue}{Then, we need to calculate how the number of these operations grows} with respect to input size. \textcolor{myblue}{Many mathematicians confirm that} \textit{asymptotic notation} \textcolor{myblue}{is necessary to express these} growth rates. \textcolor{myblue}{However, there remains debate about the most accurate methods for analyzing algorithms} with multiple variables \textcolor{myblue}{while maintaining practical relevance}. \textcolor{myblue}{Finally, we should consider the trade-offs between time complexity and space complexity} when implementing these algorithms in practice.
}
& \textbf{Similar 1:} The evaluation of mathematical proofs \textcolor{mygreen}{requires a systematic approach}. \textcolor{mygreen}{First, we must identify the basic assumptions} within the theorem. \textcolor{mygreen}{Then, we need to calculate how the logical inferences build} with respect to axiom use. \textcolor{mygreen}{Many mathematicians confirm that} \textit{formal notation} \textcolor{mygreen}{is necessary to express these} logical steps. \textcolor{mygreen}{However, there remains debate about the most accurate methods for constructing proofs} with multiple lemmas \textcolor{mygreen}{while maintaining intuitive clarity}. \textcolor{mygreen}{Finally, we should consider the trade-offs between proof elegance and proof length} when presenting these theorems in practice.
& 0.4163 & 0.6696 & 0.8386 & 0.6415 & High \\
\cmidrule(lr){2-7}
& \textbf{Similar 2:} The formulation of statistical models \textcolor{mygreen}{requires a systematic approach}. \textcolor{mygreen}{First, we must identify the basic variables} within the dataset. \textcolor{mygreen}{Then, we need to calculate how the relationships between these variables change} with respect to sample size. \textcolor{mygreen}{Many mathematicians confirm that} \textit{probabilistic notation} \textcolor{mygreen}{is necessary to express these} statistical relationships. \textcolor{mygreen}{However, there remains debate about the most accurate methods for building models} with multiple parameters \textcolor{mygreen}{while maintaining interpretability}. \textcolor{mygreen}{Finally, we should consider the trade-offs between model accuracy and model complexity} when applying these techniques in practice.
& 0.4289 & 0.7161 & 0.8474 & 0.6642 & High \\
\cmidrule(lr){2-7}
& \textbf{Similar 3:} The development of optimization algorithms \textcolor{mygreen}{requires a systematic approach}. \textcolor{mygreen}{First, we must identify the basic constraints} within the problem. \textcolor{mygreen}{Then, we need to calculate how the solution space changes} with respect to parameter values. \textcolor{mygreen}{Many mathematicians confirm that} \textit{vector notation} \textcolor{mygreen}{is necessary to express these} feasible regions. \textcolor{mygreen}{However, there remains debate about the most accurate methods for solving problems} with multiple objectives \textcolor{mygreen}{while maintaining computational efficiency}. \textcolor{mygreen}{Finally, we should consider the trade-offs between convergence speed and solution quality} when implementing these methods in practice.
& 0.4225 & 0.7116 & 0.8422 & 0.6588 & High \\
\bottomrule
\end{tabular}%
}
\caption{RCScore for Reference Paragraph vs. Similar Comparison Paragraphs. This table illustrates RCScore quantification for paragraphs with high stylistic similarity to a reference text, with visual cues for similar concepts (\textcolor{mygreen}{green text for similar concepts}, \textcolor{myblue}{blue text for reference's key phrases}).}
\label{tab:stylemetrics_example_similar_colored}
\end{table*}

\definecolor{mygreen}{rgb}{0.1, 0.5, 0.1}
\definecolor{myred}{rgb}{0.7, 0.1, 0.1}
\definecolor{myblue}{rgb}{0.1, 0.1, 0.7}
\begin{table*}[!htbp]
\centering
\resizebox{\textwidth}{!}{%
\begin{tabular}{p{0.23\textwidth} p{0.47\textwidth} r r r r l}
\toprule
\textbf{Reference} & \textbf{Comparison} & \textbf{Struct} & \textbf{Lexi} & \textbf{Cohe} & \textbf{RCScore} & \textbf{Similarity} \\
\midrule
\multirow{12}{=}{
The analysis of algorithmic complexity \textcolor{myblue}{requires a systematic approach}. \textcolor{myblue}{First, we must identify the basic operations} within the algorithm. \textcolor{myblue}{Then, we need to calculate how the number of these operations grows} with respect to input size. \textcolor{myblue}{Many mathematicians confirm that} \textit{asymptotic notation} \textcolor{myblue}{is necessary to express these} growth rates. \textcolor{myblue}{However, there remains debate about the most accurate methods for analyzing algorithms} with multiple variables \textcolor{myblue}{while maintaining practical relevance}. \textcolor{myblue}{Finally, we should consider the trade-offs between time complexity and space complexity} when implementing these algorithms in practice.
} & 
\textbf{Different 1:} \textcolor{myred}{OMG! Machine learning is SOOO amazing! �� I tried a neural network yesterday and WOW - it actually worked! Sort of... �� It got like 85\% accuracy which isn't bad for my first try, right?! I think I'm totally gonna be an AI expert now! The code was pretty simple once I figured out all those weird tensor thingies. This stuff is way cooler than boring old algorithms. \#AI \#MachineLearning \#Future}
& 0.0813 & 0.0141 & 0.7298 & 0.2751 & Low \\
\cmidrule(lr){2-7}
& \textbf{Different 2:} Consider a set X with the \textcolor{myred}{discrete topology}, where every subset of X is \textcolor{myred}{open}. If X is an \textcolor{myred}{infinite set}, such as the set of natural numbers N, then X is not \textcolor{myred}{compact}. This is because the collection of all \textcolor{myred}{singleton sets} \{x\} for x in X forms an \textcolor{myred}{open cover} of X. However, no finite \textcolor{myred}{subcollection} of this \textcolor{myred}{open cover} can cover X, as each \textcolor{myred}{singleton set} only covers one point. Therefore, by the definition of \textcolor{myred}{compactness} (a space is \textcolor{myred}{compact} if every \textcolor{myred}{open cover} has a finite \textcolor{myred}{subcover}), an \textcolor{myred}{infinite set} with the \textcolor{myred}{discrete topology} is not \textcolor{myred}{compact}. This demonstrates a fundamental concept in \textcolor{myred}{point-set topology}.
& 0.2751 & 0.1428 & 0.7821 & 0.4000 & Low \\
\cmidrule(lr){2-7}
& \textbf{Different 3:} To solve this \textcolor{myred}{quadratic equation}, first, bring all terms to one side to get the form \textcolor{myred}{ax\textasciicircum 2 + bx + c = 0}. Then, identify the \textcolor{myred}{coefficients a, b, and c}. The solutions for x can be found using the \textcolor{myred}{quadratic formula: x = (-b \textpm{} sqrt(b\textasciicircum 2 - 4ac)) / (2a)}. Remember that the \textcolor{myred}{discriminant, Delta = b\textasciicircum 2 - 4ac}, determines the nature of the roots. If \textcolor{myred}{Delta > 0}, there are two distinct real roots. If \textcolor{myred}{Delta = 0}, there is one real root (a repeated root). If \textcolor{myred}{Delta < 0}, there are two complex conjugate roots. This method is a cornerstone of \textcolor{myred}{algebra} and is widely applicable in various scientific and engineering problems.
& 0.2105 & 0.1519 & 0.7953 & 0.3859 & Low \\
\bottomrule
\end{tabular}%
}
\caption{Comparison with Semantically and Stylistically Different Paragraphs, highlighting contrasting elements (\textcolor{myred}{red text}) against the reference's structure (\textcolor{myblue}{blue text for reference's key phrases}).}
\label{tab:style-metrics-different_colored}
\end{table*}

\section{Discussion}
\label{sec:discussion}

RCScore provides a crucial lens on LLM performance by assessing sensitivity to instruction style—a dimension typically absent in standard benchmarks. Our findings confirm that models exhibit significant accuracy variations and stylistic shifts in their responses (as measured by CRS) when instruction phrasing changes, even if semantic content is preserved. This variability, influenced by task complexity and decoding strategy, highlights the limitations of single-formulation evaluations.

The notable positive correlation between CRS (particularly its lexical and aggregated components) and mean task accuracy suggests a deeper connection: models that maintain stylistic consistency when explaining solutions across varied prompts tend to achieve higher accuracy. This implies that robust problem-solving ability may manifest as more stable expressive patterns. While RCScore does not directly measure internal cognitive processes, the consistency it quantifies in response characteristics, especially when linked to accuracy on reasoning-intensive tasks, offers an empirically grounded proxy for the stability and potential depth of a model's approach. Thus, RCScore moves beyond simple performance metrics, offering diagnostic insights into the reliability and predictability of LLMs in diverse interaction contexts.

\section{Future Work}
\label{sec:future_work}

A key avenue for future work is to leverage RCScore, especially CRS, as a more direct indicator of robust reasoning capabilities. The hypothesis is that models demonstrating high CRS on complex problems across varied instruction styles are more likely to possess genuine, transferable reasoning skills, as opposed to relying on superficial pattern matching. The observed correlation between CRS and accuracy supports investigating RCScore as a complementary signal for assessing the depth and reliability of a model's problem-solving faculties.

Further applications include using RCScore's dimensional feedback to refine model architectures and training objectives to improve specific aspects of response consistency. Additionally, RCScore can inform optimized prompt engineering practices by identifying instruction styles that elicit not only accurate but also consistently well-formed responses. The framework's principles could also be extended to diverse generative tasks and modalities, and to cross-lingual style sensitivity analyses, ultimately fostering LLMs that are more adaptable and reliable.

\begin{table*}[ht]
\centering
\resizebox{\textwidth}{!}{%
\begin{tabular}{|l|cc|cc|cc|cc|}
\hline
\multirow{2}{*}{\textbf{Model}} & \multicolumn{2}{c|}{\textbf{AIME}} & \multicolumn{2}{c|}{\textbf{MATH-500}} & \multicolumn{2}{c|}{\textbf{GPQA-Diamond}} & \multicolumn{2}{c|}{\textbf{GSM8K}} \\
\cline{2-9}
 & \textbf{Beam} & \textbf{Greedy} & \textbf{Beam} & \textbf{Greedy} & \textbf{Beam} & \textbf{Greedy} & \textbf{Beam} & \textbf{Greedy} \\
\hline
Gemma 3-4B & 6.7 & \textbf{10.8} & \textbf{65.7} & 65.4 & \textbf{3.5} & 2.8 & \textbf{88.6} & 85.7 \\
Gemma 3-12B & \textbf{22.5} & 3.3 & \textbf{72.1} & 52.6 & \textbf{5.4} & 5.2 & \textbf{92.0} & 90.3 \\
Gemma 3-27B & 27.5 & 27.5 & \textbf{76.9} & 76.9 & \textbf{6.5} & 5.9 & 94.4 & \textbf{94.4} \\
\hline
LLaMA 3.2-3B & 5.0 & \textbf{11.7} & 41.1 & \textbf{43.0} & \textbf{4.1} & 3.7 & 76.2 & \textbf{79.2} \\
LLaMA 3.1-8B & 6.7 & \textbf{7.5} & 43.8 & \textbf{44.2} & 3.1 & \textbf{3.8} & 83.2 & \textbf{84.9} \\
LLaMA 3.3-70B & 23.3 & \textbf{26.7} & 63.5 & \textbf{63.6} & 6.5 & \textbf{7.4} & 90.7 & \textbf{93.4} \\
\hline
Qwen 2.5-3B & 4.2 & \textbf{8.3} & 55.0 & \textbf{57.0} & \textbf{2.9} & 2.3 & 82.5 & \textbf{84.6} \\
Qwen 2.5-7B & 11.7 & \textbf{15.9} & 62.6 & \textbf{65.8} & \textbf{4.4} & 2.1 & 86.0 & \textbf{89.5} \\
Qwen 2.5-32B & 15.0 & \textbf{16.7} & 69.1 & \textbf{70.9} & 3.4 & \textbf{5.7} & 92.6 & \textbf{92.7} \\
Qwen 2.5-72B & \textbf{22.5} & 21.7 & 70.7 & \textbf{71.1} & \textbf{5.3} & 5.1 & 92.7 & \textbf{92.9} \\
\hline
\textbf{Average} & 14.5 & \textbf{15.0} & 62.1 & \textbf{61.1} & 4.5 & 4.4 & 87.9 & \textbf{88.8} \\
\hline
\end{tabular}%
}
\caption{Average accuracy comparison between Beam Search (temperature = 1.0) and Greedy Search (temperature = 0.0) across four benchmarks. Each value represents the average of S1-S4 instruction style scores. Bold numbers indicate the better-performing decoding strategy for each model and benchmark.}
\label{tab:beam_vs_greedy_comparison}
\end{table*}

\end{document}